\documentclass[]{template}

\usepackage{multicol}
\usepackage[utf8]{inputenc}             
\usepackage[T1]{fontenc}                
\usepackage{url}                        
\usepackage{booktabs}                   
\usepackage{multirow}
\usepackage{colortbl}
\usepackage{multicol}
\usepackage{amsfonts}                   
\usepackage{nicefrac}                   
\usepackage{microtype}                  
\usepackage[dvipsnames]{xcolor}         

\usepackage{latexsym}

\usepackage{graphicx}
\usepackage{float}
\usepackage{subcaption}
\usepackage{wrapfig}
\usepackage{lipsum}

\usepackage{bm}

\usepackage{tabularx} 
\usepackage{ragged2e} 
\newcolumntype{L}{>{\RaggedRight\hangafter=1\hangindent=0em}X}

\usepackage{enumitem}

\usepackage{amsmath}
\usepackage{amssymb}
\usepackage{mathtools}
\usepackage{amsthm}

\setboolean{logo}{true}    

\usepackage{algorithm}      
\usepackage{algorithmic}    

\hypersetup{
    colorlinks=true,
    linkcolor=red,
    citecolor=Cerulean,
    filecolor=magenta,      
    urlcolor=magenta,
}

\usepackage[capitalize,noabbrev]{cleveref}
\crefname{section}{§}{§§}
\Crefname{section}{§}{§§}

\usepackage{calligra}
\DeclareMathAlphabet{\mathcalligra}{T1}{calligra}{m}{n}

\usepackage{pifont}

\theoremstyle{plain}

\theoremstyle{definition}

\theoremstyle{remark}

\renewcommand{\paragraph}[1]{\vspace{1mm}\noindent\textbf{#1}}

\usepackage[most]{tcolorbox}
\tcbset{
  promptbox/.style={
    top=10pt,
    colback=lightgray!20,
    colframe=Black,
    colbacktitle=NavyBlue,
    enhanced,
    center,
    attach boxed title to top center={yshift=-0.1in,xshift=0.0in},
    boxed title style={boxrule=0pt,colframe=white,},
  }
}
\newtcolorbox{promptbox}[2][]{promptbox, title=#2,#1}
\tcbset{
  takeawaybox/.style={
    top=10pt,
    colback=lightgray!20,
    colframe=Black,
    colbacktitle=BurntOrange,
    enhanced,
    center,
    attach boxed title to top center={yshift=-0.1in,xshift=0.0in},
    boxed title style={boxrule=0pt,colframe=white,},
  }
}
\newtcolorbox{takeawaybox}[2][]{takeawaybox, title=#2,#1}
\tcbset{
  observationbox/.style={
    top=10pt,
    colback=lightgray!20,
    colframe=Black,
    colbacktitle=YellowGreen,
    enhanced,
    center,
    attach boxed title to top center={yshift=-0.1in,xshift=0.0in},
    boxed title style={boxrule=0pt,colframe=white,},
  }
}
\newtcolorbox{observationbox}[2][]{observationbox, title=#2,#1}

\usepackage{xspace}

\newcommand\blfootnote[1]{%
  \begingroup
  \renewcommand\thefootnote{}\footnote{#1}%
  \addtocounter{footnote}{-1}%
  \endgroup
}

\usepackage{CJK}
\usepackage[table]{xcolor}
\usepackage{listings}
\usepackage{xcolor}
\usepackage{tcolorbox}
\tcbuselibrary{listings, breakable, skins}

\definecolor{deepblue}{rgb}{0,0,0.5}
\definecolor{lightblue}{rgb}{0.9,0.9,1}
\definecolor{lightgray}{rgb}{0.95,0.95,0.95}
\definecolor{darkgreen}{rgb}{0,0.4,0}
\definecolor{darkred}{rgb}{0.6,0,0}
\definecolor{lightgray}{RGB}{245,245,245}
\definecolor{mygreen}{RGB}{50,100,50} 
\definecolor{bggreen}{RGB}{220,240,220} 
\definecolor{deepblue}{RGB}{0, 51, 102}   
\definecolor{lightblue}{RGB}{173, 216, 230}  
\definecolor{oceanblue}{RGB}{0, 102, 204}   
\definecolor{skyblue}{RGB}{135, 206, 250}   
\definecolor{iceblue}{RGB}{173, 216, 230}   
\definecolor{darkice}{RGB}{65, 105, 225}    
\definecolor{techblue}{RGB}{51, 153, 255}   
\definecolor{whitebg}{RGB}{245, 255, 255}   
\definecolor{darkgreen}{rgb}{0,0.4,0}     
\definecolor{darkred}{rgb}{0.6,0,0}       
\definecolor{purple}{rgb}{0.5,0,0.5}      
\definecolor{brown}{rgb}{0.6,0.3,0}       
\definecolor{darkorange}{rgb}{0.8,0.4,0}  

\lstdefinestyle{mystyle}{
    backgroundcolor=\color{lightgray!30},
    basicstyle=\ttfamily\footnotesize,
    breakatwhitespace=false,
    breaklines=true,
    captionpos=b,
    keepspaces=true,
    showspaces=false,
    showstringspaces=false,
    showtabs=false,
    tabsize=2,
    frame=single,
    framerule=0.5pt,
    framesep=3pt,
    xleftmargin=10pt,
    xrightmargin=10pt,
    numbers=left,
    numberstyle=\tiny\color{gray},
    numbersep=5pt,
    keywordstyle=\color{deepblue},
    commentstyle=\color{darkgreen},
    stringstyle=\color{darkred},
    morekeywords={Analyze, Design, End, Action, Observation},
    morecomment=[l][\color{darkgreen}]{\#},
}
\lstdefinestyle{cleanstyle}{
    backgroundcolor=\color{lightblue!10},  
    basicstyle=\ttfamily\small,
    breaklines=true,
    breakatwhitespace=false,
    keepspaces=true,  
    showspaces=false,
    showstringspaces=false,
    showtabs=false,
    tabsize=2,
    frame=none,        
    numbers=none,      
    aboveskip=0pt,     
    belowskip=0pt,     
    xleftmargin=0pt,
    xrightmargin=0pt,
    framexleftmargin=0pt,
    framexrightmargin=0pt,
    breakindent=0pt,        
    breakautoindent=false,  
}
\title{TL-GRPO: Turn-Level RL for Reasoning-Guided Iterative Optimization}

\author[1,2]{Peiji Li}
\author[2,3]{Linyang Li}
\author[1]{Handa Sun}
\author[1]{Wenjin Mai}
\author[2]{Yongkang Chen}
\author[2]{Xiaozhe Li}
\author[2]{Yue Shen}
\author[1,2]{Yichuan Ma}
\author[2]{Yiliu Sun}
\author[2]{Jiaxi Cao}
\author[2]{Zhishu He}
\author[1]{Bo Wang}
\author[1]{Xiaoqing Zheng}
\author[1]{Zhaori Bi}
\author[1]{Xipeng Qiu}
\author[2]{Qipeng Guo}
\author[2]{Kai Chen}
\author[2,3]{Dahua Lin}

\affil[1]{Fudan University}
\affil[2]{Shanghai AI Laboratory}
\affil[3]{The Chinese University of Hong Kong}

\begin{abstract}
Large language models have demonstrated strong reasoning capabilities in complex tasks through tool integration, which is typically framed as a Markov Decision Process and optimized with trajectory-level RL algorithms such as GRPO. However, a common class of reasoning tasks—iterative optimization—presents distinct challenges: the agent interacts with the same underlying environment state across turns, and the value of a trajectory is determined by the best turn-level reward rather than cumulative returns. Existing GRPO-based methods cannot perform fine-grained, turn-level optimization in such settings, while black-box optimization methods discard prior knowledge and reasoning capabilities. To address this gap, we propose Turn-Level GRPO (TL-GRPO), a lightweight  RL algorithm that performs turn-level group sampling for fine-grained optimization. We evaluate TL-GRPO on analog circuit sizing (ACS), a challenging scientific optimization task requiring multiple simulations and domain expertise. Results show that TL-GRPO outperforms standard GRPO and Bayesian optimization methods across various specifications. Furthermore, our 30B model trained with TL-GRPO achieves state-of-the-art performance on ACS tasks under same simulation budge, demonstrating both strong generalization and practical utility.
\end{abstract}

\begin{document}

\blfootnote{$\dagger$ Corresponding authors: Linyang Li (lilinyang@pjlab.org.cn)}

\maketitle

\begin{multicols}{2}
\section{Introduction}

Large language models (LLMs) have demonstrated remarkable general reasoning capabilities~\cite{o1, deepseek-r1, qwen3}. By integrating external tools, LLM-based agents can solve many challenging and complex tasks through Tool-Integrated Reasoning(TIR)~\cite{searchr1,torl}.

\begin{figure}[H]
    \vspace{-2pt}  
  \begin{center}
    \centerline{\includegraphics[width=0.96\columnwidth]{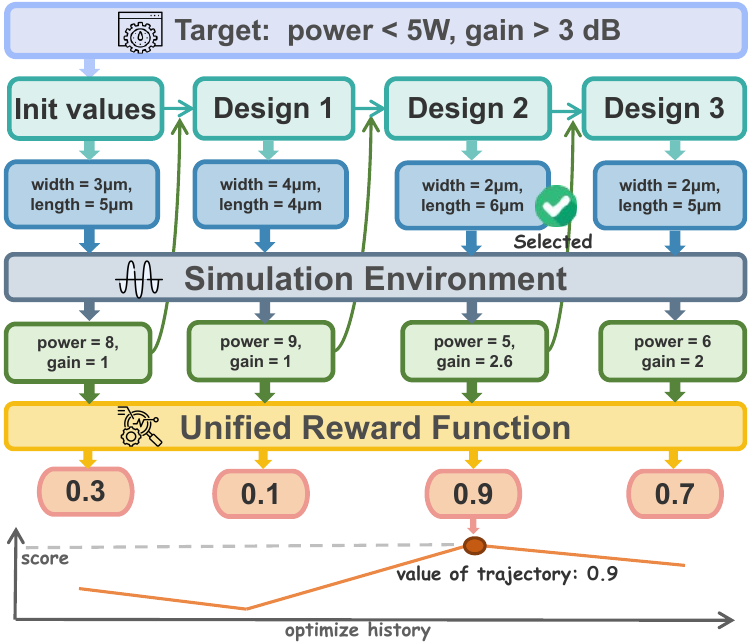}}
\caption{Illustration of an iterative optimization task (demonstrated with analog circuit sizing). Given a netlist with a set of targets, the agent iteratively proposes new designs (actions) and receives observations of the fixed environment (simulation script). In this process, each turn yields an independent reward score from a unified reward function, and the value of the entire trajectory is determined by the best design achieving the highest reward among all turns.}
    \label{fig:iteroptim}
  \end{center}
\end{figure}

Most tool-integrated reasoning tasks (e.g., search/code agent) can be formulated as partially observable Markov decision processes (POMDPs) and optimized using RL algorithms like GRPO~\cite{dsmath}. However, a distinct class of tasks—\textbf{iterative optimization}, including hyperparameter tuning, analog circuit sizing, and chemical design—exhibits fundamentally different characteristics. As illustrated in Figure~\ref{fig:iteroptim}, the agent interacts with \textbf{a fixed environment state} (e.g., a simulation script for a specific circuit task) whose intrinsic properties do not change across multiple turns. Each action receives an observation and a state-independent reward, and the value of an entire trajectory is determined by the best reward achieved in any single turn, rather than cumulative returns. These characteristics define a single-state POMDP (detailed in Sec.~\ref{sec:prelim}) , which presents distinct challenges for standard RL methods that rely on cumulative credit assignment.

\nocite{optbench}

Currently, there are three main approaches to address iterative optimization tasks:
\begin{itemize}

\vspace{-12pt}

\setlength{\itemsep}{6pt}     
\setlength{\parskip}{0pt}     
\setlength{\parsep}{0pt}      

    \item \textbf{Black-box optimization} methods (e.g., Bayesian optimization) can solve such tasks but ignores reasoning process and prior knowledge. They rely on extensive sampling, which is computationally expensive, and lack generalization across different tasks.

    

    \item \textbf{LLM-based agents} approach optimization via designed workflows \cite{alphaevolve, openevolve}, but their performance remains bounded by the base model's inherent abilities and workflow heuristics, without benefiting from specialized task training.
    
    \item \textbf{RL methods for LLMs} include single-turn RL and multi-turn tool-integrated RL. Single-turn RL cannot effectively utilize historical optimization information. And current multi-turn RL approaches, largely based on GRPO, often depend on sparse, trajectory-level/outcome-based rewards and cannot perform fine-grained optimization of each reasoning turn.
    
\vspace{-5pt} 
\end{itemize}

To address these limitations, we propose \textbf{Turn-Level GRPO (TL-GRPO)} for reasoning-guided iterative optimization. We modify the rollout strategy of GRPO: while vanilla GRPO samples multiple full trajectories for each query, TL-GRPO starts with sampling only one full trajectory until interaction completion. Then, for each tool-call turn in the trajectory, we split the trajectory at the beginning of that turn and perform turn-level group sampling while preserving history as input context. Due to the invariant environment state in iterative optimization, a single reward function can uniformly evaluate outcomes across all turns. Crucially, our algorithm introduces no additional sampling cost compared to standard GRPO.

We evaluate TL-GRPO on \textbf{Analog Circuit Sizing (ACS)}, a classic iterative optimization task with practical significance. We collect 12 professional analog circuit tasks and build a simulation environment using Cadence. By randomizing circuit initial values and target specifications, we automatically synthesize abundant query data without human annotation. We use Qwen3-30B-Instruct \cite{qwen3} to perform our RL experiments. Experimental results demonstrate that TL-GRPO outperforms standard GRPO and Bayesian optimization on both in-domain and out-of-domain evaluation sets, achieving state-of-the-art performance.

In summary, our contributions are as follows:
\begin{enumerate}
\vspace{-6pt} 
\setlength{\itemsep}{6pt}     
\setlength{\parskip}{0pt}     
\setlength{\parsep}{0pt} 

    
    \item We propose \textbf{TL-GRPO}, an RL algorithm that enables fine-grained optimization of LLM reasoning process through turn-level group sampling and a unified verifiable reward function, without introducing additional sampling cost.

    \item We formalize the unique characteristics of iterative optimization tasks and demonstrate how TL-GRPO effectively performs history-conditioned advantage estimation at each turn.
    
    \item Our 30B model trained with TL-GRPO on ACS tasks achieves state-of-the-art performance with strong generalization, providing a foundation for future LLM applications in electronic design automation (EDA).
\end{enumerate}
\section{Related Work}
\subsection{RL for LLM Reasoning}

Reinforcement learning has played a pivotal role in enhancing LLM reasoning capabilities. Algorithms such as ~\cite{ppo}, DPO~\cite{dpo}, GRPO~\cite{dsmath}, DAPO~\cite{dapo}, GSPO~\cite{gspo}, and CISPO~\cite{minimaxm1} are employed for LLM RL training to perform sequence-level preference optimization. Since GRPO relies solely on the outcome reward, many research works~\cite{synlogic,enigmata,internbootcamp,crossrlvr,interns1,go,sketch,npengine} based on the Reinforcement Learning with Verifiable Rewards (RLVR) paradigm have trained models on reasoning data from various domains to improve LLM reasoning abilities. Given that LLMs typically adopt the Chain-of-Thought~\cite{cot} reasoning paradigm, one line of work focuses on step-level supervision of fine-grained reasoning steps \cite{stepdpo,mctsdpo,fastmcts,dsmath,stepgrpo}. However, as noted by \cite{deepseek-r1}, process reward models (PRMs) require additional training or heuristic design and are susceptible to reward hacking~\cite{gao2023scaling}, while step-wise value estimation often relies on additional inference overhead (e.g., Monte Carlo sampling). Consequently, the mainstream RLVR paradigm remains dominated by GRPO methods with outcome-based rewards.

\subsection{RL for Tool-Integrated Reasoning}
Recent approaches have extended RL training to enable LLM agents to acquire multi-turn Tool-Integrated Reasoning (TIR) capabilities. A surge of research \cite{torl,toolrl,otcpo,retool,autotir,stepgrpo} employs trajectory-level RL algorithms for this purpose. These works typically incorporate tool-execution outcomes to provide intermediate reward signals, which are aggregated into one outcome reward, making it difficult to identify the contribution of individual tool-calling turns. To address this, some research has begun to explore more fine-grained, turn-level advantage estimation. Methods such as SPARL~\cite{sparl}, MT-PPO~\cite{mtppo}, and GTPO~\cite{gtpo} acquire intermediate rewards through various heuristics, while GiGPO~\cite{gigpo} calculates turn-level advantage via anchor-state grouping and Tree-GRPO~\cite{treegrpo} transforms trajectory-level signals into process supervision through back-propagation. However, these methods are designed for general POMDPs and still face challenges in precise credit assignment. For iterative optimization tasks, \cite{kevin} propose a multi-turn RL approach for generating CUDA kernels that normalizes rewards from all turns in GRPO for advantage estimation, which ignores the impact of historical context.


\subsection{Analog Circuit Sizing}

Analog circuit sizing (ACS) is a classical optimization task in which engineers apply domain knowledge and experience to tune parameters such that a circuit meets specified objectives (e.g., gain, power). Prior machine learning approaches have largely framed ACS as a black-box optimization problem, employing methods like Bayesian optimization \cite{parallelbayes,batchbayes} that lack generalization across different circuit topologies. Recent work has explored LLM-based solutions, including multi-agent frameworks \cite{adollm,anaflow,eesizer, Artisan} and supervised fine-tuning \cite{easysize}, to improve generalization across ACS tasks. However, these methods either depend heavily on the base model's inherent capabilities or require costly human-annotated data for training. To overcome these limitations, we formulate ACS as a \textbf{reasoning-guided} iterative optimization task and introduce a turn-level RL paradigm that enables the LLM to learn from experience and generalize effectively across diverse circuit specifications.
\section{Preliminaries}
\label{sec:prelim}

\subsection{Problem Formulation}

In standard multi-turn Tool-Integrated Reasoning tasks, the interaction process is typically formalized as a Partially Observable Markov Decision Process (POMDP) \cite{landscapeagenticrl}, defined by the tuple:
\begin{equation}
 \langle \mathcal{S}, \mathcal{A}, \mathcal{P}, \mathcal{R}, \gamma, \mathcal{O} \rangle,
\end{equation}
where the agent receives an observation $o_t = O(s_t)$ based on the state $s_t \in \mathcal{S}$, takes an action $a_t \in \mathcal{A}_{\text{text}} \cup \mathcal{A}_{\text{tool\_call}}$, and the environment transitions according to $\mathcal{P}(s_{t+1} \mid s_t, a_t)$. The objective of POMDP is to maximize the discounted cumulative reward:
\begin{equation}
  \mathcal{J}_{\text{POMDP}}(\theta) = \mathbb{E}_{\tau \sim \pi_\theta} \left[ \sum_{t=0}^{T-1} \gamma^{t} R(s_t, a_t) \right], \quad 0 < \gamma < 1.
\label{eq:pomdp_objective}
\end{equation}

However, \textbf{iterative optimization} tasks constitute a distinct class where the agent aims to find the best possible solution within a fixed budget of turns. Formally, such tasks can be modeled as a \textbf{single-state POMDP}. Similar to multi-armed bandits, the agent interacts with a fixed environment that represents the evaluator of the optimization task. The state $s$, corresponding to the evaluator's intrinsic properties, remains unchanged throughout the interaction. This leads to a trivial state transition:
\begin{equation}
\mathcal{P}(s' \mid s, a) = \delta(s' = s), \quad \forall a \in \mathcal{A},
\label{eq:single_state_transition}
\end{equation}
where $\delta(\cdot)$ is the Dirac delta function. At each turn $t$, the agent proposes a candidate design $\mathbf{x}_t$ (via action $a_t$) and receives an observation $o_t = O(s, a_t)$ (e.g., performance metrics) and a reward $r_t = R(a_t)$ that depends only on the action. The underlying optimization objective is to find:
\begin{equation}
\mathbf{x}^* = \arg\max_{\mathbf{x} \in \mathcal{X}} R(\mathbf{x}), \quad \mathbf{x} \in \mathbb{R}^d,
\label{eq:optim}
\end{equation}
where $R: \mathcal{X} \rightarrow \mathbb{R}$ is a scalar function that balances multiple objectives, and $\mathcal{X} \subseteq \mathbb{R}^d$ is a $d$-dimensional design space.

Two fundamental properties characterize iterative optimization and distinguish it from standard POMDP formulations:

\paragraph{Single-State Structure with Action-Dependent Feedback.} Due to the invariant environment state, observations and rewards vary solely based on the action $a_t$ taken. The agent can utilize the entire history of interactions $h_t = (a_0, o_0, \dots, a_{t-1}, o_{t-1})$ to inform its decisions, enabling it to reason about the design space and improve proposals over successive turns. Crucially, the same deterministic reward function $R$ uniformly evaluates every turn (e.g., \mbox{Figure of Merit} in ACS), making each reward $r_t = R(a_t)$ a self-contained measure of optimization quality.

\paragraph{Non-Cumulative, Maximum-Reward Objective.} The goal is not to maximize the sum of rewards, but to achieve the highest possible reward in any single turn. Therefore, for iterative optimization, the value of a trajectory $\tau = (a_0, o_0, r_0, \dots, a_{T-1}, o_{T-1}, r_{T-1})$ is:
\begin{align}
G(\tau) = \max_{t \in [0, T-1]} r_t = \max_{t \in [0, T-1]} R(a_t).
\label{eq:traj_value}
\end{align}
Consequently, the optimization objective for an iterative optimization agent is:
\begin{equation}
\mathcal{J}_{\text{Opt}}(\theta) = \mathbb{E}_{\tau \sim \pi_\theta} \left[\, \max_{t \in [0, T-1]} R(a_t) \,\right].
\label{eq:objective}
\end{equation}

This objective presents dual challenges for standard RL methods. First, it defies the cumulative credit assignment inherent in Eq.~\ref{eq:pomdp_objective}. Second, it yields sparse trajectory-level signals that fail to distinguish the quality of individual turns, hindering fine-grained optimization. Our Turn-Level GRPO addresses these challenges by performing turn-level group rollouts to estimate per-turn advantages, which optimizes each turn independently and eliminates the need for complex credit assignment across the sequence.

\begin{figure*}[ht]
    \centering
    \includegraphics[width=\linewidth]{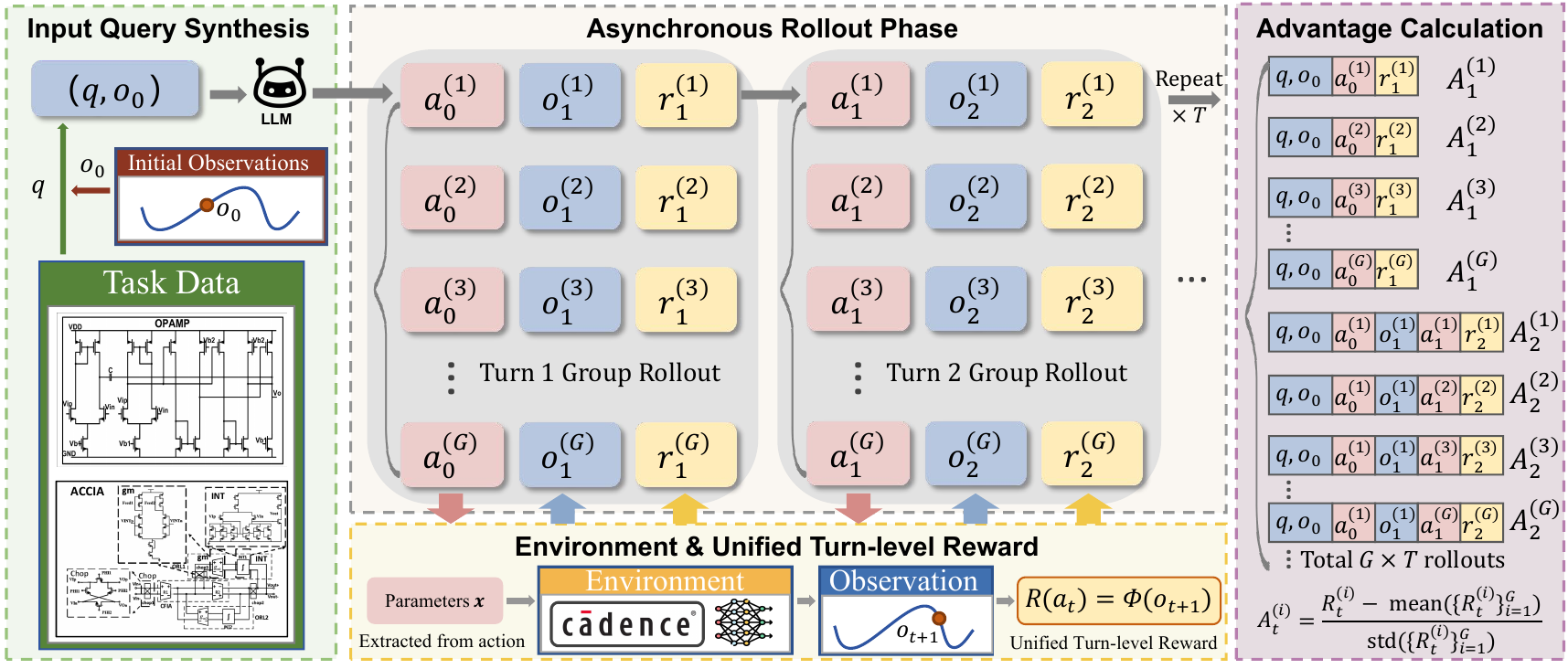}
\caption{Overview of TL-GRPO. For an input query with initial observations, the LLM agent asynchronously performs turn-level group sampling. Each sampled action $a
^{(i)}_t \in \mathcal{A}_{\text{text}} \cup \mathcal{A}_{\text{tool\_call}}$ receives an observation with a turn-level reward, and a group-relative advantage is subsequently estimated. All turn-level rollouts are then collected into a single batch for policy gradient updates.}

    \label{fig:tlgrpo}
\end{figure*}

\subsection{Group Relative Policy Optimization}

Group Relative Policy Optimization is a reinforcement learning method widely adopted to enhance the reasoning capabilities of LLMs. For each input query $x$, GRPO samples a group of $G$ responses $\{y_1, y_2, \ldots, y_G\}$ from old policy $\pi_{\theta_{\text{old}}}$ and optimizes the following objective:
\begin{equation}
\begin{aligned}
& \mathcal{J}_\text{GRPO}(\theta) =  \mathbb{E}_{ x, \{y_i\}^{G}_{i=1} \sim \pi_{\theta_{\text{old}}}} 
\Bigg[ \frac{1}{G} \sum_{i=1}^{G}\frac{1}{|y_i|}  \sum_{t=1}^{|y_i|} \min \\
&\left(w_{i,t}(\theta) {A}_{i,t}, \, \mathrm{clip} \left( w_{i,t}(\theta), 1 - {\varepsilon}, 1 + {\varepsilon}\right) {A}_{i,t} \right) \\
& - \beta \mathbb{D}_{\text{KL}}[ \pi_\theta \,\|\, \pi_{\text{ref}}] \Bigg],
\label{eq:grpo}
\end{aligned}
\end{equation}
where $w_{i,t}(\theta)=\frac{\pi_\theta(y_{i,t} \mid x, y_{i,<t})}{\pi_{\theta_{\text{old}}}(y_{i,t} \mid x, y_{i,<t})}$ is the token-level importance sampling ratio, $\varepsilon$ is the clipping parameter, $\beta$ controls the KL regularization against the reference policy $\pi_{\text{ref}}$.

Given one group of trajectory-level rewards $\{R^{\text{traj}}_i\}_{i=1}^G$, it computes a normalized advantage for entire $i$-th response:
\begin{equation}
A_{i,t} = A_{i} =  \frac{R^{\text{traj}}_i - \text{mean}(\{R^{\text{traj}}_i\}_{i=1}^G)}{\text{std}(\{R^{\text{traj}}_i\}_{i=1}^G)}.
\label{eq:grpo-adv}
\end{equation}
As shown in Eq.~\ref{eq:grpo-adv}, the advantage $A_{i,t}$ is identical for all tokens within the same trajectory, applying a uniform, trajectory-level credit assignment. This design is effective in standard reasoning tasks, as test-time-scaling is shown to be beneficial for reasoning~\cite{o1}. However, for iterative optimization tasks—where the value of a trajectory is defined by the maximum turn reward rather than a final outcome—this coarse-grained credit assignment provides no distinction between high- and low-quality turns. Consequently, applying vanilla GRPO to iterative optimization often results in noisy updates, unstable training, and suboptimal policy performance.
\section{Method: Turn-Level GRPO}

To address the limitations of directly applying vanilla GRPO to iterative optimization tasks, we propose Turn-Level Group Relative Policy Optimization (TL-GRPO). Our approach centers on two key strategies: (i) turn-level group sampling and (ii) unified verifiable turn rewards. With these modifications, TL-GRPO achieves fine-grained, turn-level preference optimization without introducing additional sampling cost compared to vanilla GRPO. An overview of our algorithm is illustrated in Figure \ref{fig:tlgrpo}.

\subsection{Turn-level Group Sampling}
\label{sec:turnlevelsampling}

To enable turn-level value estimation without introducing additional computational overhead, we depart from the trajectory-level group rollout strategy used in vanilla GRPO. In TL-GRPO, for each input query $q$, we begin with sampling \textbf{only one full trajectory} from the old policy $\pi_{\theta_{\text{old}}}$, denoted as $\tau = (o_0, a_0, o_1, a_1, ..., o_{T-1}, a_{T-1})$.

We then split this trajectory at each turn, obtaining $T$ distinct history contexts:
\begin{align}
    h_0 &= (q, o_0), \nonumber \\
    h_1 &= (q, o_0, a_0, o_1), \nonumber \\
    h_2 &= (q, o_0, a_0, o_1, a_1, o_2), \nonumber \\
    &\quad \vdots \nonumber \\
    h_{T-1} &= (q, o_0, a_0, o_1, a_1, ..., o_{T-1}).
\end{align}

For each history context $h_t$, we perform a group rollout by sampling $G$ actions from the current policy:
\begin{align}
     \{a_t^{(i)}\}_{i=1}^{G} &\sim \pi_\theta \bigl( a_t^{(i)} \mid h_t \bigr) = \pi_\theta \bigl( a_t^{(i)} \mid (q, o_0, a_0, \dots, o_t) \bigr), \nonumber \\
    &\text{where} \quad a_t \in \mathcal{A}_{\text{text}} \cup \mathcal{A}_{\text{tool\_call}}.
\end{align}
This allows us to compute a group-relative advantage for each turn using the same normalized advantage estimation as in vanilla GRPO, but applied at turn level.

\paragraph{Inference Budget}
A key advantage of this design is that TL-GRPO maintains theoretically the same inference cost as vanilla GRPO. Let the expected number of turns in a trajectory be $T$ and the group size be $G$. Both methods require $G \times T$ turn-level rollouts per query. This design avoids the substantial per-turn sampling overhead concerns noted in prior work \cite{gigpo,mtppo}, as TL-GRPO does not perform group sampling over  full trajectories, but only performs group sampling at each turn.

\subsection{Unified Verifiable Turn-level Function}
\label{sec:unifiedverifiable}
The second core component of TL-GRPO is the use of a unified, verifiable reward function for all turns. As described in Sec.~\ref{sec:turnlevelsampling}, each action $a_t$ results in an observation $o_{t+1}$, which comprises a vector of multiple performance metrics (e.g., gain, bandwidth, power). The key step is to map this multi-objective vector into a scalar score that quantifies the overall quality of the proposed parameters $a_t$:
\begin{equation}
R(a_t) = \Phi(o_{t+1}), \quad \Phi: \mathbb{R}^{|o_t|} \rightarrow \mathbb{R}
\label{eq:unified_reward}
\end{equation}
where $\Phi$ is a known, deterministic function that combines the various performance metrics into a single reward. This unified reward provides a direct, turn-level signal for optimization. In our scenario, no temporal discounting or cumulative credit assignment is required, as the final objective (Eq.~\ref{eq:objective}) depends only on the maximum reward achieved across turns, meaning each turn-level reward independently contends for being the best. This design ensures a consistent optimization target across all turns and eliminates the need for complex temporal credit assignment, making it particularly suitable for iterative optimization.

\subsection{Learning Objective of TL-GRPO}
Building upon the turn-level group sampling and unified reward function described above, we now formalize the complete optimization objective of TL-GRPO:
\begin{align}\label{eq:tlgrpo}
   & \mathcal{J}_\text{TL-GRPO}(\theta)  =  \mathbb{E}_{ q, \tau \sim \pi_{\theta_{\text{old}}}}  \Bigg[ \frac{1}{G \times T} \sum_{t=1}^T \sum_{i=1}^G \nonumber \\
   & \min\Bigg( \frac{\pi_\theta(a_t^{(i)} | h_t)}{\pi_{\theta_{old}}(a_t^{(i)} | h_t)} {A}_t^{(i)}, \nonumber  \\
   & \mathrm{clip} \left( \frac{\pi_\theta(a_t^{(i)} | h_t)}{\pi_{\theta_{old}}(a_t^{(i)} | h_t)}, 1 - \varepsilon, 1+ \varepsilon \right) {A}_t^{(i)} \Bigg)\\
   & - \beta \mathbb{D}_{\text{KL}}[ \pi_\theta \,\|\, \pi_{\text{ref}}]\Bigg], \nonumber 
\end{align}
where each group of turn-level sampling outputs $\{a_t^{(i)}\}_{i=1}^{G} = \{a_t^{(1)}, a_t^{(2)}, \ldots, a_{t}^{(G)}\}$ is sampled from old policy $\pi_{\theta_{old}}$ from every history context $h_t$. The KL regularization term  aligns with standard GRPO~\cite{dsmath} but can be omitted in practice if stability permits~\cite{dapo}.

For each turn $t$, the group-relative advantage ${A}_t^{(i)}$ for the $i$-th rollout $a_t^{(i)}$ is computed by normalizing the rewards within the turn-level sampling group:

\begin{equation}\label{eq:GRPO-Adv}
    {A}_t^{(i)} = \frac{{R}_{t}^{(i)} - \mathrm{mean}(\{{R}_{t}^{(i)}\}_{i=1}^{G})}{\mathrm{std}(\{{R}_{t}^{(i)}\}_{i=1}^{G})}.
\end{equation} 

Algorithm~\ref{alg:tlgrpo} summarizes the complete TL-GRPO procedure. In practice, we collect all turn-level rollouts into a single batch for policy gradient updates. Through this design, TL-GRPO provides fine-grained, turn-level preference optimization for multi-turn iterative optimization tasks. Since $a_t \in \mathcal{A}_{\text{text}} \cup \mathcal{A}_{\text{tool\_call}}$, this method offers turn-level supervision for both the reasoning process and the tool-calling strategy. Note that although the turn-level reward function is unified across turns, the advantage estimate ${A}_t^{(i)}$ is computed conditioned on the specific history context $h_t$. Consequently, by optimizing the conditional policy $\pi_\theta ( a_t^{(i)} \mid h_t )$, the agent learns from historical interaction experience, which goes beyond merely fitting a static response distribution and enables the policy to leverage past context for learning from experience\cite{eraofexperience}.

\newcommand{\COMMENTLLAMA}[1]{{\textcolor[HTML]{A9A9A9} {$\triangleright$ {#1}\\}}}
\newcommand{\COMMENTLIGHTGRAY}[1]{\hfill{\textcolor[HTML]{A9A9A9} {$\triangleright$  {#1}\\}}}

\begin{algorithm}[H]
\caption{Turn-level GRPO}
\label{alg:tlgrpo}
\begin{algorithmic}
\STATE {\bfseries Input:} Initial policy $\pi_{\theta_{\text{old}}}$, unified reward function $R$, dataset $\mathcal{D}$, group size $G$, max turns $T$, clip parameters $\varepsilon$
\STATE {\bfseries Output: }$\pi_{\theta}$
\FOR{each training iteration}
\STATE Update old policy: $\pi_{\theta_{\text{old}}} \gets \pi_{\theta}$
\STATE Sample a batch of input queries $\mathcal{Q}_b \sim \mathcal{D}$
\STATE \COMMENTLLAMA{Asynchronous rollout phase}
\FOR{each query $q \in \mathcal{Q}_b$ }
\STATE Generate one full trajectory  
\STATE $\quad \tau = (o_0, a_0, o_1, \dots, o_{T-1}, a_{T-1}) \sim \pi_{\theta_{\text{old}}}(\cdot \mid q)$
\STATE Split $\tau$ at each turn to obtain history contexts:
\STATE $\quad \{h_t\}_{t=0}^{T-1}$ where $h_t = (q, o_0, a_0, \dots, o_t)$
\FOR{$t=0$ {\bfseries to} $T-1$} 
\STATE Sampling $G$ actions $\{a_t^{(i)}\}_{i=1}^{G} \sim \pi_{\theta_{\text{old}}}(\cdot \mid h_t)$
\STATE Compute turn-level rewards $\{r_t^{(i)}\}_{i=1}^{G}$
\STATE Calculate turn-level advantage ${A}_t^{(i)}$  (Eq.~\ref{eq:GRPO-Adv})
\ENDFOR 
\ENDFOR 
\STATE \COMMENTLIGHTGRAY{Collect all turn-level rollouts into one batch}
\STATE Update policy model $\pi_{\theta}$ by maximizing objective $\mathcal{J}_\text{TL-GRPO}(\theta)$ (Eq.~\ref{eq:tlgrpo})
\ENDFOR

\end{algorithmic}
\vspace{-2pt}  
\end{algorithm}
\section{Experiments}

To evaluate the effectiveness of TL-GRPO, we conduct experiments on Analog Circuit Sizing (ACS), which is a classic, real-world iterative optimization problem with significant scientific and engineering implications. We compare our algorithm against traditional black-box optimization (Bayesian Optimization), single-turn GRPO, and trajectory-level GRPO, demonstrating its advantages in both performance and generalization.

\subsection{Experimental Setup}
\subsubsection{Environments for Analog Circuit Sizing}

We implement our environment based on InternBootcamp~\cite{internbootcamp}. To ensure task diversity, we collect 12 real-world analog circuit sizing tasks with practical significance, including amplifiers, data converters and others, with different netlists and PDK (Process Design Kit). Specific performance targets (specifications) are set to reflect production requirements. The objective is to tune designated parameters in the circuit netlist to meet these multi-objective performance targets. See Appendix \ref{apd:acs}
for details.

The primary challenges in this setting are: (1) simulation tools (e.g., Cadence Spectre) require complex environment configuration; (2) circuit simulation is computationally intensive and time-consuming. To address these, we build a distributed master-worker framework that decouples model inference from tool execution. The LLM agent runs in the training or inference loops while simulation tools are deployed as independent worker services registered to a central master node. During both training and inference, tool calls are made via requests to the master, which schedules and executes the simulations asynchronously. This architecture isolates the agent inference loop from the heavy simulation environment, improving efficiency and scalability.

\subsubsection{Data Synthesis}

To effectively validate our algorithm within a tractable experimental framework, we first simplify the optimization problem based on domain expertise. We restrict each tunable parameter to a physically reasonable search range (e.g., $w_{1} \in [0.4, 2] \, \text{\textmu m} )$ and limit the maximum tool-calling turns to \( T = 5 \). This design reflects a practical balance between exploration cost and convergence likelihood in real-world iterative optimization, allowing us to focus on evaluating the core capability of turn-level reasoning optimization.

Each analog circuit task initially corresponds to a single query with a fixed set of tunable parameters and production-realistic targets, which is insufficient for training. To create a diverse and scalable dataset from this simplified setup, we introduce two modifications under the guidance of human domain experts: (1) \textbf{randomizing the initial values} of all tunable parameters within the predefined reasonable range for each query, and (2) \textbf{applying controlled random offsets} to original target specification values, creating distinct optimization targets within practical design margins. Given that a single task can involve dozens of parameters and multiple objectives, these variations yield substantial diversity, enabling the synthesis of a large volume of unique training queries without manual annotation.

Moreover, our prompt requires the LLM agent to first explicitly articulate its reasoning process within \texttt{<Analyze> </Analyze>} tags before calling the simulation tool and to conclude the interaction with \texttt{<Design End>}. This structured format clearly separates reasoning process from tool execution, aligning with the reasoning-guided iterative optimization paradigm. Additionally, it allows the reward function to differentiate between agent-initiated termination and external truncation. Example prompts and tool descriptions are provided in Appendix \ref{apd:dataexample}.

\subsubsection{Unified Turn Reward Design}
\label{sec:rewarddesign}

As formalized in Section~\ref{sec:unifiedverifiable}, TL-GRPO employs a unified verifiable reward function $\Phi$ that maps the observation $o_t$ (the tool response) into a scalar turn-level reward $R_t = \Phi(o_t)$. For the Analog Circuit Sizing task, the observation $o_t$ contains the values of multiple performance metrics (e.g., gain, bandwidth, phase margin, power consumption) resulting from the proposed circuit parameters.

Let $\mathcal{M} = \{m_1, m_2, \dots, m_M\}$ denote the set of $M$ optimization objectives. Each objective $m_j$ has a current value $v_j$ (extracted from tool response $o_t$) and a target specification $s_j$ that defines the desired performance. The specifications can be of three types: \textbf{lower-bound} (e.g., minimum gain), \textbf{upper-bound} (e.g., maximum power), and \textbf{range} (e.g., output voltage range). For each objective, we define a normalized score $p_j(v_j, s_j) \in [0, 1]$ that quantifies how well $v_j$ meets $s_j$, using threshold parameters $\tau_j^{\ell}$ and $\tau_j^{u}$ to create smooth transitions between acceptable and unacceptable performance (see Appendix~\ref{apd:reward_details} for details). 

The overall performance reward $P_t \in [0, 1]$ is computed as the geometric mean of all per-objective scores:
\begin{equation}
    P_t = \left( \prod_{j=1}^{M} p_j(v_j, s_j) \right)^{1/M}
    \label{eq:performance_reward}
\end{equation}
The geometric mean ensures that the agent must balance all objectives simultaneously, as a low score in any single objective will significantly reduce $P_t$. This unified turn-level reward function serves like a Figure of Merit (FoM) in analog circuit design, but normalized to $[0,1]$ for consistent evaluation across different tasks.

Additionally, we introduce a format penalty $F_t \in [-1, 0]$ during training that penalizes violations of the required interaction format, such as missing \texttt{<Analyze>} or \texttt{<Design End>} tags, or exceeding the maximum tool-calling turns $T$. In evaluation, for fair comparison, we do not apply format penalty. The final reward is:
\begin{align}
    \Phi_t =
    \begin{cases}
    \max\!\big(0,\; \min(1,\; P_t + F_t) \big), &\text{for training} \\
    P_t, & \text{for eval}
    \end{cases}
    \label{eq:final_reward}
\end{align}

where $\Phi_t \in [-1, 0]$. This design provides a consistent, verifiable, and fine-grained turn-level reward signal that directly aligns with the multi-objective nature of the ACS task while encouraging proper interaction behavior.

\begin{table*}[htb]
\footnotesize
\caption{In-domain evaluation results (8 trained tasks). \textit{ST-iter} refers to the single-turn iterative evaluation protocol (Sec.~\ref{sec:setups}). }
\centering
\label{tab:indomain}
\begin{tabular}{lcccccccccc}
\toprule
Method & Eval & ACCIA & ADC1 &  ADC2 & ADC3 & ADC4 & ADCL & OTA &  SimLDO & \textbf{Avg.} \\ 
\midrule
Bayesian Optimization &  & 0.51 & 0.83 & 0.23 & 0.78 & 0.86 & 0.41 & 0.69 & 0.20 & 0.56\\
\midrule
Qwen3-235B-Instruct-2507 & Multi-turn & 0.29 & 0.76 & 0.06 & 0.75 & 0.77 & 0.55 & 0.66 & 0.01 & 0.48 \\
 & ST-iter & 0.33 & 0.88 & 0.07 & 0.85 & 0.87 & 0.60 & 0.66 & 0.01 & 0.53 \\
DeepSeek-V3.2 (no-think) & Multi-turn & 0.23 & 0.89 & 0.31 & 0.81 & 0.89 & 0.79 & 0.63 & 0.46 & 0.63  \\
 & ST-iter & 0.24 & 0.93 & 0.34 & 0.83 & 0.88 & 0.60 & 0.67 & 0.50 & 0.62\\
DeepSeek-V3.2 (thinking) & Multi-turn & 0.26 & 0.85 & 0.31 & 0.78 & 0.81 & 0.78 & 0.65 & 0.55 & 0.62 \\
 & ST-iter & 0.25 & 0.89 & 0.33 & 0.80 & 0.74 & 0.53 & 0.66 & 0.65 & 0.61 \\
\midrule
Qwen3-30B-Instruct-2507 & Multi-turn & 0.20 & 0.65 & 0.15 & 0.71 & 0.76 & 0.50 & 0.59 & 0.01 & 0.45 \\
 & ST-iter & 0.18 & 0.79 & 0.21 & 0.75 & 0.83 & 0.39 & 0.65 & 0.00 & 0.47 \\
\rowcolor{lightgray}
+ Trajectory-Level GRPO & Multi-turn & 0.69 & 0.98 & 0.74 & 0.99 & 0.98 & 0.92 & 0.87 & 0.61 & 0.85\\
\rowcolor{lightgray}
+ Single-Turn GRPO & ST-iter & 0.68 & 0.99 & 0.61 & 0.93 & 0.97 & 0.93  & 0.93 & 0.57 & 0.83\\
\rowcolor{lightgray}
+ Turn-Level GRPO (Ours) & Multi-turn & \textbf{0.94} & \textbf{1.0} & \textbf{0.91} & \textbf{1.0} & \textbf{1.0} & \textbf{0.98} & \textbf{0.99} & \textbf{0.97} & \textbf{0.97}\\
\bottomrule
\end{tabular}
\end{table*}

\begin{table*}[htb]
\caption{Out-of-domain evaluation results (4 unseen tasks). \textit{ST-iter} refers to the single-turn iterative evaluation protocol (Sec.~\ref{sec:setups}).}
\centering
\label{tab:ood}
\begin{tabular}{lccccccc}
\toprule
Method & Eval & OPAMP1 & OPAMP2 & OPAMP3 & OPAMP4 & \textbf{Avg.} \\
\midrule
Bayesian Optimization &   & \textbf{0.36} & 0.01 & 0.08 & 0.24 & 0.17 \\

\midrule
Qwen3-235B-Instruct-2507 & Multi-turn &  0.21 & 0.04 & 0.08 & 0.19 & 0.13 \\
 & ST-iter & 0.23 & 0.17 & 0.15 & 0.24 & 0.20 \\
DeepSeek-V3.2 (no-think) & Multi-turn & 0.25 & 0.23 & 0.24 & 0.30 & 0.26 \\
 & ST-iter & 0.22 & 0.26 & 0.24 & 0.24 & 0.24 \\
DeepSeek-V3.2 (thinking) & Multi-turn & 0.22 & 0.23 & 0.26 & 0.26 & 0.24\\
 & ST-iter & 0.20 & 0.28  & \textbf{0.32} & 0.23 & 0.26 \\
\midrule
Qwen3-30B-Instruct-2507 & Multi-turn & 0.14 & 0.01 & 0.05 & 0.16 & 0.09\\
 & ST-iter & 0.14 & 0.04 & 0.06 & 0.15 & 0.10\\
 \rowcolor{lightgray}
+ Trajectory-Level GRPO & Multi-turn & 0.15 & 0.18 & 0.20 & 0.17 & 0.18 \\
\rowcolor{lightgray}
+ Single-Turn GRPO & ST-iter & 0.20 & 0.15 & 0.26 & 0.25 & 0.21 \\
\rowcolor{lightgray}
+ Turn-Level GRPO (Ours) & Multi-turn &0.20 & \textbf{0.32} & 0.31 & \textbf{0.58} & \textbf{0.35}\\

\bottomrule
\end{tabular}
\end{table*}

\begin{figure*}[t]
  \centering
  \begin{subfigure}{0.33\textwidth}
    \centering
    \includegraphics[width=\linewidth]{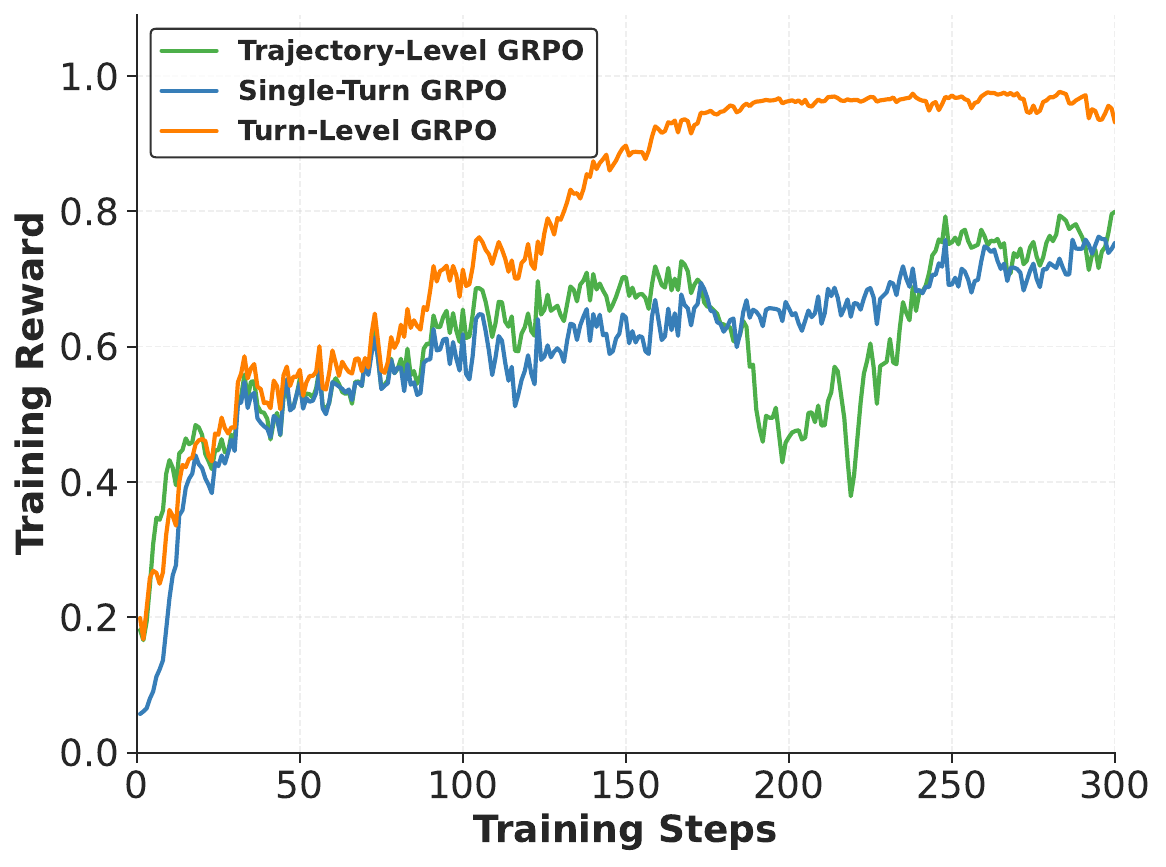}
    \caption{Reward score.}
    \label{fig:reward}
  \end{subfigure}
  \hfill
  \begin{subfigure}{0.32\textwidth}
    \centering
    \includegraphics[width=\linewidth]{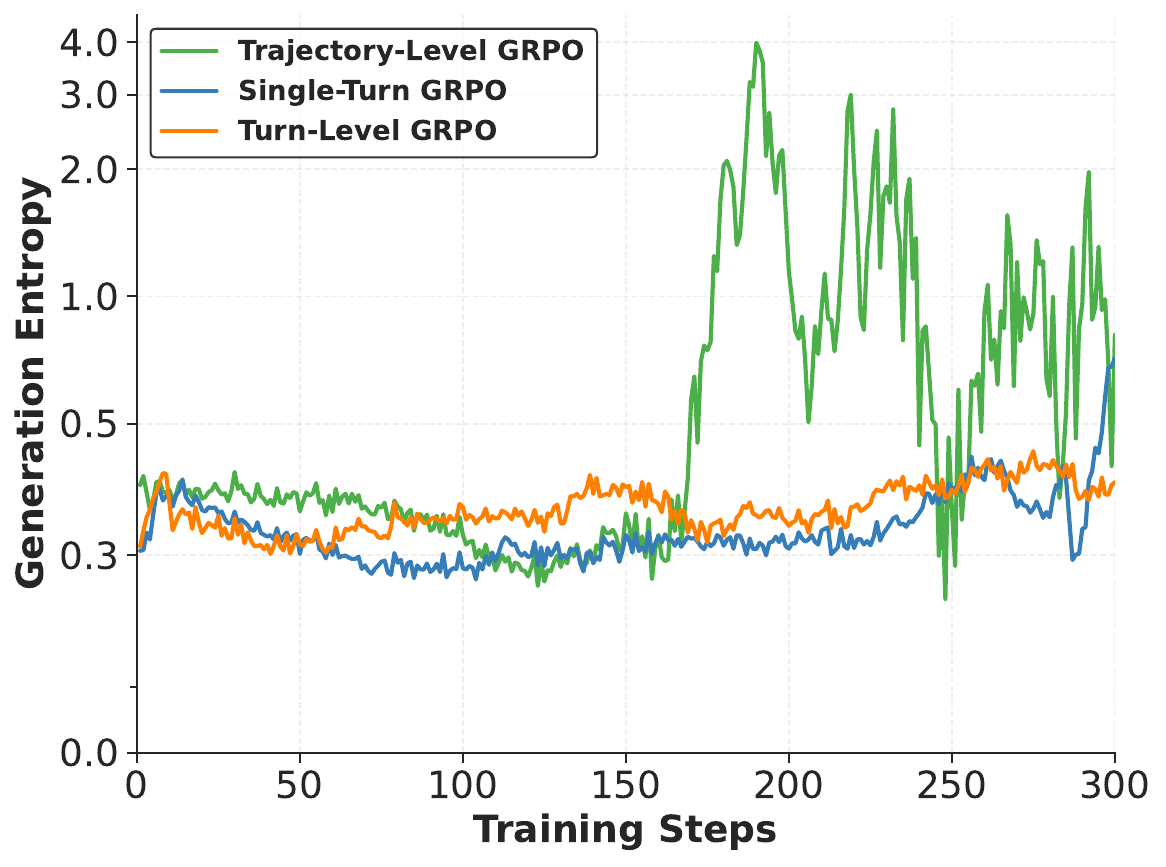}
    \caption{Generation entropy of actor model.}
    \label{fig:entropy}
  \end{subfigure}
  \hfill
  \begin{subfigure}{0.33\textwidth}
    \centering
    \includegraphics[width=\linewidth]{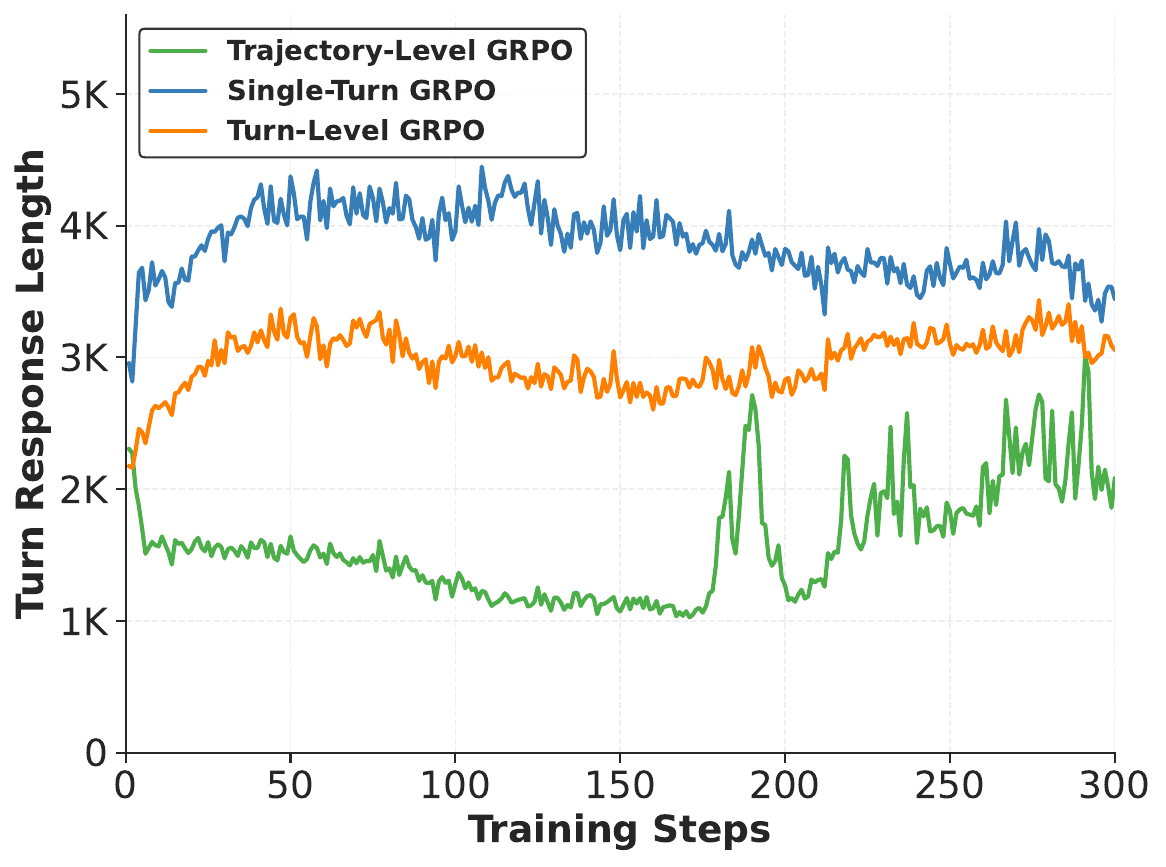}
    \caption{Average turn response length.}
    \label{fig:responselen}
  \end{subfigure}

  \caption{Comparison of training dynamics for the three RL algorithms: Trajectory-Level GRPO, Single-Turn GRPO, and Turn-Level GRPO. (a) Training reward score over steps. (b) Actor model generation entropy. (c) Average turn response length.}
  \label{fig:dynamics}
  \vspace{-8pt}
\end{figure*}


\subsubsection{Benchmarks and Training Setups}
\label{sec:setups}

The 12 circuit tasks are partitioned into training and evaluation sets. 8 tasks with distinct netlists are used for training and in-domain evaluation, while 4 Op-Amp tasks with different PDK libraries are held out for out-of-domain evaluation. We synthesize 10K training queries and generate 100 distinct queries per task for evaluation, resulting in 800 in-domain and 400 out-of-domain instances.

We compare three training paradigms: (1) TL-GRPO (multi-turn rollout, turn-level optimization), (2) Trajectory-Level GRPO (trajectory-level optimization, advantage calculated with maximum reward across turns), and (3) Single-Turn GRPO (single-turn optimization without history).  We adopt Qwen3-30B-A3B-Instruct as the base model, with batch size 32, rollout group size $G=8$, and a maximum of $T=5$ turns per query for multi-turn settings. Other training setup details are demonstrated in Appendix \ref{apd:trainingsetup}.

During evaluation, we compare the three RL paradigms mentioned above, Bayesian Optimization \cite{bayespython}, and strong open-source LLMs (Qwen3-235B-Instruct and DeepSeek-V3.2)\footnote{Due to data confidentiality demands from collaborators, we are unable to use APIs from closed-source models for evaluation.}. The evaluation setting aligns with training: each query starts from a random initial simulation point and allows a maximum of 5 tool-call turns, and we report the best candidate score as final score. Two evaluation protocols for LLMs are used: (i) Multi-turn evaluation, which aligns with TL-GRPO and Trajectory-level GRPO that optimize multi-turn reasoning; (ii) Single-turn iterative evaluation, where an LLM iteratively proposes a new design based only on the original query and the most recent simulation result, aligning with Single-Turn GRPO training.

\subsection{Main Results}

Table~\ref{tab:indomain} and \ref{tab:ood} present the evaluation performance of TL-GRPO and all baseline methods under the same simulation budget (a maximum of 5 tool-call turns). We report results from the best checkpoint for each method. Examples of trained model's outputs are provided in Appendix~\ref{apd:dataexample}.

On in-domain evaluation set (Table~\ref{tab:indomain}), TL-GRPO achieves the highest scores across all 8 trained ACS tasks, with an average score of 0.97. This demonstrates that our method enables the 30B model to master the trained tasks effectively. Notably, it converges faster and to a higher final performance than both Trajectory-Level and Single-Turn GRPO.

The out-of-domain (OOD) evaluation (Table~\ref{tab:ood}) holds greater practical significance, as real-world optimization tasks like ACS are computationally expensive, limiting the feasibility of extensive sampling or iteration for new problems, thus demanding strong generalization. Here, TL-GRPO again achieves the highest average score (0.35), surpassing Bayesian Optimization (0.17) and powerful LLMs including Qwen3-235B-Instruct (0.20) and DeepSeek-V3.2 (0.26).

Our analysis yields three key findings:

1.  Compared to Bayesian Optimization, TL-GRPO demonstrates the value of \textbf{reasoning-guided iterative optimization}, where prior knowledge and reasoning enable better generalization to unseen tasks.

2.  Compared to Trajectory-Level GRPO, TL-GRPO's superior performance validates that \textbf{turn-level} fine-grained  optimization strengthens reasoning capability more effectively than trajectory-level, outcome-based rewards, which introduce noise in credit assignment

3.  Compared to Single-Turn GRPO, TL-GRPO highlights the importance of \textbf{historical context}; by conditioning each turn on the full interaction history, TL-GRPO naturally \textit{learns from experience} across iterations, leading to more informed design proposals


\subsection{Analysis of Training Dynamics}

\Cref{fig:dynamics} compares the training reward, generation entropy, and average response length across TL-GRPO, Trajectory-Level GRPO, and Single-Turn GRPO, effectively serving as an analysis that isolates the impact of turn-level optimization in our method relative to standard GRPO baselines.

The reward curve in \cref{fig:reward} demonstrates that TL-GRPO achieves faster convergence and reaches higher final rewards compared to both baselines. Furthermore, as shown in \cref{fig:responselen,fig:entropy}, TL-GRPO maintains stable response lengths and well-behaved generation entropy throughout training, despite the challenge of optimizing sequences that can potentially reach 35k tokens (due to a maximum of 15k-token prompt and up to 5 $\times$ 4k-token turn response). The entropy does not collapse to low values—which would indicate reduced exploration—nor does it exhibit sharp fluctuations, suggesting stable and controlled policy updates. Example reasoning traces from our TL-GRPO trained model are shown in \cref{fig:example_response_turn_1,fig:example_response_turn_2,fig:example_response_turn_3,fig:example_response_turn_4,fig:example_response_turn_5}. For a detailed turn-level analysis of optimization dynamics, see Appendix~\ref{apd:turn_analysis}.

We attribute TL-GRPO's effectiveness to its fine-grained optimization strategy: by independently optimizing each turn within a trajectory, it provides the most detailed learning signal possible. In contrast, Trajectory-Level GRPO assigns a uniform advantage estimate to all tokens in one sequence, which introduces significant noise and leads to the observed instability and reward oscillations. 
\section{Conclusion}

We introduce Turn-Level Group Relative Policy Optimization (TL-GRPO), an efficient multi-turn tool-integrated RL algorithm for reasoning-guided iterative optimization tasks. By employing a turn-level group sampling strategy and a unified verifiable turn-level reward design, TL-GRPO achieves fine-grained, turn-level preference optimization without incurring additional sampling cost compared to vanilla GRPO. Experiments on analog circuit sizing demonstrate that TL-GRPO outperforms both trajectory-level and single-turn GRPO, and exhibits strong generalization to unseen circuit specifications, achieving state-of-the-art performance on our benchmark. We believe our work offers a practical approach for enhancing LLM reasoning in optimization tasks and paves the way for future LLM applications in electronic design automation.

\section{Limitations}

Our work has several limitations that point to directions for future research:

\textbf{Resource constraints.} While we compare TL-GRPO with Bayesian optimization (BO) on analog circuit sizing, both methods operate under a limited simulation budget. This constraint is due to the computational expense of circuit simulation, where a single evaluation can take minutes to hours for complex circuits. Although we selected tasks with relatively fast simulations, training still requires significant computational resources. The comparison with BO under this limited budget remains valid, as we restrict the parameter search ranges to physically reasonable intervals, making the optimization tractable. Future work with increased simulation budgets would further validate our approach.

\textbf{Turn-level reward design.} While our unified, verifiable reward function effectively normalizes multiple objectives and provides a preference signal across actions, it may not be theoretically optimal. Furthermore, it may not fully capture the value of exploratory actions that maylead to better designs in later turns. Future work could explore more sophisticated reward calculation methods that better balance immediate performance with long-term exploration.

\textbf{Dataset scale and diversity.} We evaluated TL-GRPO on 12 analog circuit tasks, which, while diverse, represent only a subset of real-world optimization problems. The high commercial value and confidentiality of industrial circuit designs limit data availability. Collecting more tasks would improve generalization. Additionally, we plan to extend our evaluation to other scientific optimization domains to further validate the broad applicability of our method.

\clearpage
\bibliographystyle{plain}
\bibliography{refs}

@article{o1,
  author       = {Aaron Jaech and
                  Adam Kalai and
                  Adam Lerer and
                  Adam Richardson and
                  Ahmed El{-}Kishky and
                  Aiden Low and
                  Alec Helyar and
                  Aleksander Madry and
                  Alex Beutel and
                  Alex Carney and
                  Alex Iftimie and
                  Alex Karpenko and
                  Alex Tachard Passos and
                  Alexander Neitz and
                  Alexander Prokofiev and
                  Alexander Wei and
                  Allison Tam and
                  Ally Bennett and
                  Ananya Kumar and
                  Andre Saraiva and
                  Andrea Vallone and
                  Andrew Duberstein and
                  Andrew Kondrich and
                  Andrey Mishchenko and
                  Andy Applebaum and
                  Angela Jiang and
                  Ashvin Nair and
                  Barret Zoph and
                  Behrooz Ghorbani and
                  Ben Rossen and
                  Benjamin Sokolowsky and
                  Boaz Barak and
                  Bob McGrew and
                  Borys Minaiev and
                  Botao Hao and
                  Bowen Baker and
                  Brandon Houghton and
                  Brandon McKinzie and
                  Brydon Eastman and
                  Camillo Lugaresi and
                  Cary Bassin and
                  Cary Hudson and
                  Chak Ming Li and
                  Charles de Bourcy and
                  Chelsea Voss and
                  Chen Shen and
                  Chong Zhang and
                  Chris Koch and
                  Chris Orsinger and
                  Christopher Hesse and
                  Claudia Fischer and
                  Clive Chan and
                  Dan Roberts and
                  Daniel Kappler and
                  Daniel Levy and
                  Daniel Selsam and
                  David Dohan and
                  David Farhi and
                  David Mely and
                  David Robinson and
                  Dimitris Tsipras and
                  Doug Li and
                  Dragos Oprica and
                  Eben Freeman and
                  Eddie Zhang and
                  Edmund Wong and
                  Elizabeth Proehl and
                  Enoch Cheung and
                  Eric Mitchell and
                  Eric Wallace and
                  Erik Ritter and
                  Evan Mays and
                  Fan Wang and
                  Felipe Petroski Such and
                  Filippo Raso and
                  Florencia Leoni and
                  Foivos Tsimpourlas and
                  Francis Song and
                  Fred von Lohmann and
                  Freddie Sulit and
                  Geoff Salmon and
                  Giambattista Parascandolo and
                  Gildas Chabot and
                  Grace Zhao and
                  Greg Brockman and
                  Guillaume Leclerc and
                  Hadi Salman and
                  Haiming Bao and
                  Hao Sheng and
                  Hart Andrin and
                  Hessam Bagherinezhad and
                  Hongyu Ren and
                  Hunter Lightman and
                  Hyung Won Chung and
                  Ian Kivlichan and
                  Ian O'Connell and
                  Ian Osband and
                  Ignasi Clavera Gilaberte and
                  Ilge Akkaya},
  title        = {OpenAI o1 System Card},
  journal      = {CoRR},
  volume       = {abs/2412.16720},
  year         = {2024},
  url          = {https://doi.org/10.48550/arXiv.2412.16720},
  doi          = {10.48550/ARXIV.2412.16720},
  eprinttype    = {arXiv},
  eprint       = {2412.16720},
  bibsource    = {dblp computer science bibliography, https://dblp.org}
}

@article{deepseek-r1,
  author       = {DeepSeek{-}AI},
  title        = {DeepSeek-R1: Incentivizing Reasoning Capability in LLMs via Reinforcement
                  Learning},
  journal      = {CoRR},
  volume       = {abs/2501.12948},
  year         = {2025},
  url          = {https://doi.org/10.48550/arXiv.2501.12948},
  doi          = {10.48550/ARXIV.2501.12948},
  eprinttype    = {arXiv},
  eprint       = {2501.12948},
  bibsource    = {dblp computer science bibliography, https://dblp.org}
}

@article{qwen3,
  author       = {An Yang and
                  Anfeng Li and
                  Baosong Yang and
                  Beichen Zhang and
                  Binyuan Hui and
                  Bo Zheng and
                  Bowen Yu and
                  Chang Gao and
                  Chengen Huang and
                  Chenxu Lv and
                  Chujie Zheng and
                  Dayiheng Liu and
                  Fan Zhou and
                  Fei Huang and
                  Feng Hu and
                  Hao Ge and
                  Haoran Wei and
                  Huan Lin and
                  Jialong Tang and
                  Jian Yang and
                  Jianhong Tu and
                  Jianwei Zhang and
                  Jian Yang and
                  Jiaxi Yang and
                  Jingren Zhou and
                  Junyang Lin and
                  Kai Dang and
                  Keqin Bao and
                  Kexin Yang and
                  Le Yu and
                  Lianghao Deng and
                  Mei Li and
                  Mingfeng Xue and
                  Mingze Li and
                  Pei Zhang and
                  Peng Wang and
                  Qin Zhu and
                  Rui Men and
                  Ruize Gao and
                  Shixuan Liu and
                  Shuang Luo and
                  Tianhao Li and
                  Tianyi Tang and
                  Wenbiao Yin and
                  Xingzhang Ren and
                  Xinyu Wang and
                  Xinyu Zhang and
                  Xuancheng Ren and
                  Yang Fan and
                  Yang Su and
                  Yichang Zhang and
                  Yinger Zhang and
                  Yu Wan and
                  Yuqiong Liu and
                  Zekun Wang and
                  Zeyu Cui and
                  Zhenru Zhang and
                  Zhipeng Zhou and
                  Zihan Qiu},
  title        = {Qwen3 Technical Report},
  journal      = {CoRR},
  volume       = {abs/2505.09388},
  year         = {2025},
  url          = {https://doi.org/10.48550/arXiv.2505.09388},
  doi          = {10.48550/ARXIV.2505.09388},
  eprinttype    = {arXiv},
  eprint       = {2505.09388},
  bibsource    = {dblp computer science bibliography, https://dblp.org}
}

@article{interns1,
  author       = {Lei Bai and
                  Zhongrui Cai and
                  Yuhang Cao and
                  Maosong Cao and
                  Weihan Cao and
                  Chiyu Chen and
                  Haojiong Chen and
                  Kai Chen and
                  Pengcheng Chen and
                  Ying Chen and
                  Yongkang Chen and
                  Yu Cheng and
                  Pei Chu and
                  Tao Chu and
                  Erfei Cui and
                  Ganqu Cui and
                  Long Cui and
                  Ziyun Cui and
                  Nianchen Deng and
                  Ning Ding and
                  Nanqing Dong and
                  Peijie Dong and
                  Shihan Dou and
                  Sinan Du and
                  Haodong Duan and
                  Caihua Fan and
                  Ben Gao and
                  Changjiang Gao and
                  Jianfei Gao and
                  Songyang Gao and
                  Yang Gao and
                  Zhangwei Gao and
                  Jiaye Ge and
                  Qiming Ge and
                  Lixin Gu and
                  Yuzhe Gu and
                  Aijia Guo and
                  Qipeng Guo and
                  Xu Guo and
                  Conghui He and
                  Junjun He and
                  Yili Hong and
                  Siyuan Hou and
                  Caiyu Hu and
                  Hanglei Hu and
                  Jucheng Hu and
                  Ming Hu and
                  Zhouqi Hua and
                  Haian Huang and
                  Junhao Huang and
                  Xu Huang and
                  Zixian Huang and
                  Zhe Jiang and
                  Lingkai Kong and
                  Linyang Li and
                  Peiji Li and
                  Pengze Li and
                  Shuaibin Li and
                  Tianbin Li and
                  Wei Li and
                  Yuqiang Li and
                  Dahua Lin and
                  Junyao Lin and
                  Tianyi Lin and
                  Zhishan Lin and
                  Hongwei Liu and
                  Jiangning Liu and
                  Jiyao Liu and
                  Junnan Liu and
                  Kai Liu and
                  Kaiwen Liu and
                  Kuikun Liu and
                  Shichun Liu and
                  Shudong Liu and
                  Wei Liu and
                  Xinyao Liu and
                  Yuhong Liu and
                  Zhan Liu and
                  Yinquan Lu and
                  Haijun Lv and
                  Hongxia Lv and
                  Huijie Lv and
                  Qitan Lv and
                  Ying Lv and
                  Chengqi Lyu and
                  Chenglong Ma and
                  Jianpeng Ma and
                  Ren Ma and
                  Runmin Ma and
                  Runyuan Ma and
                  Xinzhu Ma and
                  Yichuan Ma and
                  Zihan Ma and
                  Sixuan Mi and
                  Junzhi Ning and
                  Wenchang Ning and
                  Xinle Pang and
                  Jiahui Peng and
                  Runyu Peng and
                  Yu Qiao},
  title        = {Intern-S1: {A} Scientific Multimodal Foundation Model},
  journal      = {CoRR},
  volume       = {abs/2508.15763},
  year         = {2025},
  url          = {https://doi.org/10.48550/arXiv.2508.15763},
  doi          = {10.48550/ARXIV.2508.15763},
  eprinttype    = {arXiv},
  eprint       = {2508.15763},
  bibsource    = {dblp computer science bibliography, https://dblp.org}
}

@article{internbootcamp,
  author       = {Peiji Li and
                  Jiasheng Ye and
                  Yongkang Chen and
                  Yichuan Ma and
                  Zijie Yu and
                  Kedi Chen and
                  Ganqu Cui and
                  Haozhan Li and
                  Jiacheng Chen and
                  Chengqi Lyu and
                  Wenwei Zhang and
                  Linyang Li and
                  Qipeng Guo and
                  Dahua Lin and
                  Bowen Zhou and
                  Kai Chen},
  title        = {InternBootcamp Technical Report: Boosting {LLM} Reasoning with Verifiable
                  Task Scaling},
  journal      = {CoRR},
  volume       = {abs/2508.08636},
  year         = {2025},
  url          = {https://doi.org/10.48550/arXiv.2508.08636},
  doi          = {10.48550/ARXIV.2508.08636},
  eprinttype    = {arXiv},
  eprint       = {2508.08636},
  bibsource    = {dblp computer science bibliography, https://dblp.org}
}

@inproceedings{fastmcts,
  author       = {Peiji Li and
                  Kai Lv and
                  Yunfan Shao and
                  Yichuan Ma and
                  Linyang Li and
                  Xiaoqing Zheng and
                  Xipeng Qiu and
                  Qipeng Guo},
  editor       = {Wanxiang Che and
                  Joyce Nabende and
                  Ekaterina Shutova and
                  Mohammad Taher Pilehvar},
  title        = {FastMCTS: {A} Simple Sampling Strategy for Data Synthesis},
  booktitle    = {Proceedings of the 63rd Annual Meeting of the Association for Computational
                  Linguistics (Volume 1: Long Papers), {ACL} 2025, Vienna, Austria,
                  July 27 - August 1, 2025},
  pages        = {24405--24422},
  publisher    = {Association for Computational Linguistics},
  year         = {2025},
  url          = {https://aclanthology.org/2025.acl-long.1190/},
  bibsource    = {dblp computer science bibliography, https://dblp.org}
}

@article{landscapeagenticrl,
  author       = {Guibin Zhang and
                  Hejia Geng and
                  Xiaohang Yu and
                  Zhenfei Yin and
                  Zaibin Zhang and
                  Zelin Tan and
                  Heng Zhou and
                  Zhongzhi Li and
                  Xiangyuan Xue and
                  Yijiang Li and
                  Yifan Zhou and
                  Yang Chen and
                  Chen Zhang and
                  Yutao Fan and
                  Zihu Wang and
                  Songtao Huang and
                  Yue Liao and
                  Hongru Wang and
                  Mengyue Yang and
                  Heng Ji and
                  Michael Littman and
                  Jun Wang and
                  Shuicheng Yan and
                  Philip Torr and
                  Lei Bai},
  title        = {The Landscape of Agentic Reinforcement Learning for LLMs: {A} Survey},
  journal      = {CoRR},
  volume       = {abs/2509.02547},
  year         = {2025},
  url          = {https://doi.org/10.48550/arXiv.2509.02547},
  doi          = {10.48550/ARXIV.2509.02547},
  eprinttype    = {arXiv},
  eprint       = {2509.02547},
  bibsource    = {dblp computer science bibliography, https://dblp.org}
}

@article{searchr1,
  author       = {Bowen Jin and
                  Hansi Zeng and
                  Zhenrui Yue and
                  Dong Wang and
                  Hamed Zamani and
                  Jiawei Han},
  title        = {Search-R1: Training LLMs to Reason and Leverage Search Engines with
                  Reinforcement Learning},
  journal      = {CoRR},
  volume       = {abs/2503.09516},
  year         = {2025},
  url          = {https://doi.org/10.48550/arXiv.2503.09516},
  doi          = {10.48550/ARXIV.2503.09516},
  eprinttype    = {arXiv},
  eprint       = {2503.09516},
  bibsource    = {dblp computer science bibliography, https://dblp.org}
}

@article{dsmath,
  author       = {Zhihong Shao and
                  Peiyi Wang and
                  Qihao Zhu and
                  Runxin Xu and
                  Junxiao Song and
                  Mingchuan Zhang and
                  Y. K. Li and
                  Y. Wu and
                  Daya Guo},
  title        = {DeepSeekMath: Pushing the Limits of Mathematical Reasoning in Open
                  Language Models},
  journal      = {CoRR},
  volume       = {abs/2402.03300},
  year         = {2024},
  url          = {https://doi.org/10.48550/arXiv.2402.03300},
  doi          = {10.48550/ARXIV.2402.03300},
  eprinttype    = {arXiv},
  eprint       = {2402.03300},
  bibsource    = {dblp computer science bibliography, https://dblp.org}
}

@article{ppo,
  author       = {John Schulman and
                  Filip Wolski and
                  Prafulla Dhariwal and
                  Alec Radford and
                  Oleg Klimov},
  title        = {Proximal Policy Optimization Algorithms},
  journal      = {CoRR},
  volume       = {abs/1707.06347},
  year         = {2017},
  url          = {http://arxiv.org/abs/1707.06347},
  eprinttype    = {arXiv},
  eprint       = {1707.06347},
  bibsource    = {dblp computer science bibliography, https://dblp.org}
}

@article{dapo,
  author       = {Qiying Yu and
                  Zheng Zhang and
                  Ruofei Zhu and
                  Yufeng Yuan and
                  Xiaochen Zuo and
                  Yu Yue and
                  Tiantian Fan and
                  Gaohong Liu and
                  Lingjun Liu and
                  Xin Liu and
                  Haibin Lin and
                  Zhiqi Lin and
                  Bole Ma and
                  Guangming Sheng and
                  Yuxuan Tong and
                  Chi Zhang and
                  Mofan Zhang and
                  Wang Zhang and
                  Hang Zhu and
                  Jinhua Zhu and
                  Jiaze Chen and
                  Jiangjie Chen and
                  Chengyi Wang and
                  Hongli Yu and
                  Weinan Dai and
                  Yuxuan Song and
                  Xiangpeng Wei and
                  Hao Zhou and
                  Jingjing Liu and
                  Wei{-}Ying Ma and
                  Ya{-}Qin Zhang and
                  Lin Yan and
                  Mu Qiao and
                  Yonghui Wu and
                  Mingxuan Wang},
  title        = {{DAPO:} An Open-Source {LLM} Reinforcement Learning System at Scale},
  journal      = {CoRR},
  volume       = {abs/2503.14476},
  year         = {2025},
  url          = {https://doi.org/10.48550/arXiv.2503.14476},
  doi          = {10.48550/ARXIV.2503.14476},
  eprinttype    = {arXiv},
  eprint       = {2503.14476},
  bibsource    = {dblp computer science bibliography, https://dblp.org}
}

@article{gspo,
  title={Group Sequence Policy Optimization}, 
  author={
    Chujie Zheng and Shixuan Liu and Mingze Li and Xiong-Hui Chen and Bowen Yu and 
    Chang Gao and Kai Dang and Yuqiong Liu and Rui Men and An Yang and Jingren Zhou and 
    Junyang Lin 
  },
  journal={arXiv preprint arXiv:2507.18071},
  year={2025}
}

@article{minimaxm1,
  author       = {Aili Chen and
                  Aonian Li and
                  Bangwei Gong and
                  Binyang Jiang and
                  Bo Fei and
                  Bo Yang and
                  Boji Shan and
                  Changqing Yu and
                  Chao Wang and
                  Cheng Zhu and
                  Chengjun Xiao and
                  Chengyu Du and
                  Chi Zhang and
                  Chu Qiao and
                  Chunhao Zhang and
                  Chunhui Du and
                  Congchao Guo and
                  Da Chen and
                  Deming Ding and
                  Dianjun Sun and
                  Dong Li and
                  Enwei Jiao and
                  Haigang Zhou and
                  Haimo Zhang and
                  Han Ding and
                  Haohai Sun and
                  Haoyu Feng and
                  Huaiguang Cai and
                  Haichao Zhu and
                  Jian Sun and
                  Jiaqi Zhuang and
                  Jiaren Cai and
                  Jiayuan Song and
                  Jin Zhu and
                  Jingyang Li and
                  Jinhao Tian and
                  Jinli Liu and
                  Junhao Xu and
                  Junjie Yan and
                  Junteng Liu and
                  Junxian He and
                  Kaiyi Feng and
                  Ke Yang and
                  Kecheng Xiao and
                  Le Han and
                  Leyang Wang and
                  Lianfei Yu and
                  Liheng Feng and
                  Lin Li and
                  Lin Zheng and
                  Linge Du and
                  Lingyu Yang and
                  Lunbin Zeng and
                  Minghui Yu and
                  Mingliang Tao and
                  Mingyuan Chi and
                  Mozhi Zhang and
                  Mujie Lin and
                  Nan Hu and
                  Nongyu Di and
                  Peng Gao and
                  Pengfei Li and
                  Pengyu Zhao and
                  Qibing Ren and
                  Qidi Xu and
                  Qile Li and
                  Qin Wang and
                  Rong Tian and
                  Ruitao Leng and
                  Shaoxiang Chen and
                  Shaoyu Chen and
                  Shengmin Shi and
                  Shitong Weng and
                  Shuchang Guan and
                  Shuqi Yu and
                  Sichen Li and
                  Songquan Zhu and
                  Tengfei Li and
                  Tianchi Cai and
                  Tianrun Liang and
                  Weiyu Cheng and
                  Weize Kong and
                  Wenkai Li and
                  Xiancai Chen and
                  Xiangjun Song and
                  Xiao Luo and
                  Xiao Su and
                  Xiaobo Li and
                  Xiaodong Han and
                  Xinzhu Hou and
                  Xuan Lu and
                  Xun Zou and
                  Xuyang Shen and
                  Yan Gong and
                  Yan Ma and
                  Yang Wang and
                  Yiqi Shi and
                  Yiran Zhong and
                  Yonghong Duan},
  title        = {MiniMax-M1: Scaling Test-Time Compute Efficiently with Lightning Attention},
  journal      = {CoRR},
  volume       = {abs/2506.13585},
  year         = {2025},
  url          = {https://doi.org/10.48550/arXiv.2506.13585},
  doi          = {10.48550/ARXIV.2506.13585},
  eprinttype    = {arXiv},
  eprint       = {2506.13585},
  bibsource    = {dblp computer science bibliography, https://dblp.org}
}

@article{dpo,
  title={Direct preference optimization: Your language model is secretly a reward model},
  author={Rafailov, Rafael and Sharma, Archit and Mitchell, Eric and Manning, Christopher D and Ermon, Stefano and Finn, Chelsea},
  journal={Advances in neural information processing systems},
  volume={36},
  pages={53728--53741},
  year={2023}
}

@article{synlogic,
  author       = {Junteng Liu and
                  Yuanxiang Fan and
                  Zhuo Jiang and
                  Han Ding and
                  Yongyi Hu and
                  Chi Zhang and
                  Yiqi Shi and
                  Shitong Weng and
                  Aili Chen and
                  Shiqi Chen and
                  Yunan Huang and
                  Mozhi Zhang and
                  Pengyu Zhao and
                  Junjie Yan and
                  Junxian He},
  title        = {SynLogic: Synthesizing Verifiable Reasoning Data at Scale for Learning
                  Logical Reasoning and Beyond},
  journal      = {CoRR},
  volume       = {abs/2505.19641},
  year         = {2025},
  url          = {https://doi.org/10.48550/arXiv.2505.19641},
  doi          = {10.48550/ARXIV.2505.19641},
  eprinttype    = {arXiv},
  eprint       = {2505.19641},
  bibsource    = {dblp computer science bibliography, https://dblp.org}
}

@article{enigmata,
  author       = {Jiangjie Chen and
                  Qianyu He and
                  Siyu Yuan and
                  Aili Chen and
                  Zhicheng Cai and
                  Weinan Dai and
                  Hongli Yu and
                  Qiying Yu and
                  Xuefeng Li and
                  Jiaze Chen and
                  Hao Zhou and
                  Mingxuan Wang},
  title        = {Enigmata: Scaling Logical Reasoning in Large Language Models with
                  Synthetic Verifiable Puzzles},
  journal      = {CoRR},
  volume       = {abs/2505.19914},
  year         = {2025},
  url          = {https://doi.org/10.48550/arXiv.2505.19914},
  doi          = {10.48550/ARXIV.2505.19914},
  eprinttype    = {arXiv},
  eprint       = {2505.19914},
  bibsource    = {dblp computer science bibliography, https://dblp.org}
}

@article{crossrlvr,
  author       = {Yi Su and
                  Dian Yu and
                  Linfeng Song and
                  Juntao Li and
                  Haitao Mi and
                  Zhaopeng Tu and
                  Min Zhang and
                  Dong Yu},
  title        = {Crossing the Reward Bridge: Expanding {RL} with Verifiable Rewards
                  Across Diverse Domains},
  journal      = {CoRR},
  volume       = {abs/2503.23829},
  year         = {2025},
  url          = {https://doi.org/10.48550/arXiv.2503.23829},
  doi          = {10.48550/ARXIV.2503.23829},
  eprinttype    = {arXiv},
  eprint       = {2503.23829},
  bibsource    = {dblp computer science bibliography, https://dblp.org}
}

@article{cot,
  title={Chain-of-thought prompting elicits reasoning in large language models},
  author={Wei, Jason and Wang, Xuezhi and Schuurmans, Dale and Bosma, Maarten and Xia, Fei and Chi, Ed and Le, Quoc V and Zhou, Denny and others},
  journal={Advances in neural information processing systems},
  volume={35},
  pages={24824--24837},
  year={2022}
}

@article{stepdpo,
  author       = {Xin Lai and
                  Zhuotao Tian and
                  Yukang Chen and
                  Senqiao Yang and
                  Xiangru Peng and
                  Jiaya Jia},
  title        = {Step-DPO: Step-wise Preference Optimization for Long-chain Reasoning
                  of LLMs},
  journal      = {CoRR},
  volume       = {abs/2406.18629},
  year         = {2024},
  url          = {https://doi.org/10.48550/arXiv.2406.18629},
  doi          = {10.48550/ARXIV.2406.18629},
  eprinttype    = {arXiv},
  eprint       = {2406.18629},
  bibsource    = {dblp computer science bibliography, https://dblp.org}
}

@article{mctsdpo,
  author       = {Yuxi Xie and
                  Anirudh Goyal and
                  Wenyue Zheng and
                  Min{-}Yen Kan and
                  Timothy P. Lillicrap and
                  Kenji Kawaguchi and
                  Michael Shieh},
  title        = {Monte Carlo Tree Search Boosts Reasoning via Iterative Preference
                  Learning},
  journal      = {CoRR},
  volume       = {abs/2405.00451},
  year         = {2024},
  url          = {https://doi.org/10.48550/arXiv.2405.00451},
  doi          = {10.48550/ARXIV.2405.00451},
  eprinttype    = {arXiv},
  eprint       = {2405.00451},
  bibsource    = {dblp computer science bibliography, https://dblp.org}
}

@inproceedings{gao2023scaling,
  title={Scaling laws for reward model overoptimization},
  author={Gao, Leo and Schulman, John and Hilton, Jacob},
  booktitle={International Conference on Machine Learning},
  pages={10835--10866},
  year={2023},
  organization={PMLR}
}

@article{toolrl,
  author       = {Cheng Qian and
                  Emre Can Acikgoz and
                  Qi He and
                  Hongru Wang and
                  Xiusi Chen and
                  Dilek Hakkani{-}T{\"{u}}r and
                  Gokhan Tur and
                  Heng Ji},
  title        = {ToolRL: Reward is All Tool Learning Needs},
  journal      = {CoRR},
  volume       = {abs/2504.13958},
  year         = {2025},
  url          = {https://doi.org/10.48550/arXiv.2504.13958},
  doi          = {10.48550/ARXIV.2504.13958},
  eprinttype    = {arXiv},
  eprint       = {2504.13958},
  bibsource    = {dblp computer science bibliography, https://dblp.org}
}

@article{otcpo,
  author       = {Hongru Wang and
                  Cheng Qian and
                  Wanjun Zhong and
                  Xiusi Chen and
                  Jiahao Qiu and
                  Shijue Huang and
                  Bowen Jin and
                  Mengdi Wang and
                  Kam{-}Fai Wong and
                  Heng Ji},
  title        = {{OTC:} Optimal Tool Calls via Reinforcement Learning},
  journal      = {CoRR},
  volume       = {abs/2504.14870},
  year         = {2025},
  url          = {https://doi.org/10.48550/arXiv.2504.14870},
  doi          = {10.48550/ARXIV.2504.14870},
  eprinttype    = {arXiv},
  eprint       = {2504.14870},
  bibsource    = {dblp computer science bibliography, https://dblp.org}
}

@article{torl,
  author       = {Xuefeng Li and
                  Haoyang Zou and
                  Pengfei Liu},
  title        = {ToRL: Scaling Tool-Integrated {RL}},
  journal      = {CoRR},
  volume       = {abs/2503.23383},
  year         = {2025},
  url          = {https://doi.org/10.48550/arXiv.2503.23383},
  doi          = {10.48550/ARXIV.2503.23383},
  eprinttype    = {arXiv},
  eprint       = {2503.23383},
  bibsource    = {dblp computer science bibliography, https://dblp.org}
}

@article{retool,
  author       = {Jiazhan Feng and
                  Shijue Huang and
                  Xingwei Qu and
                  Ge Zhang and
                  Yujia Qin and
                  Baoquan Zhong and
                  Chengquan Jiang and
                  Jinxin Chi and
                  Wanjun Zhong},
  title        = {ReTool: Reinforcement Learning for Strategic Tool Use in LLMs},
  journal      = {CoRR},
  volume       = {abs/2504.11536},
  year         = {2025},
  url          = {https://doi.org/10.48550/arXiv.2504.11536},
  doi          = {10.48550/ARXIV.2504.11536},
  eprinttype    = {arXiv},
  eprint       = {2504.11536},
  bibsource    = {dblp computer science bibliography, https://dblp.org}
}

@article{autotir,
  author       = {Yifan Wei and
                  Xiaoyan Yu and
                  Yixuan Weng and
                  Tengfei Pan and
                  Angsheng Li and
                  Li Du},
  title        = {AutoTIR: Autonomous Tools Integrated Reasoning via Reinforcement Learning},
  journal      = {CoRR},
  volume       = {abs/2507.21836},
  year         = {2025},
  url          = {https://doi.org/10.48550/arXiv.2507.21836},
  doi          = {10.48550/ARXIV.2507.21836},
  eprinttype    = {arXiv},
  eprint       = {2507.21836},
  bibsource    = {dblp computer science bibliography, https://dblp.org}
}

@article{stepgrpo,
  author       = {Jingyi Zhang and
                  Jiaxing Huang and
                  Huanjin Yao and
                  Shunyu Liu and
                  Xikun Zhang and
                  Shijian Lu and
                  Dacheng Tao},
  title        = {{R1-VL:} Learning to Reason with Multimodal Large Language Models
                  via Step-wise Group Relative Policy Optimization},
  journal      = {CoRR},
  volume       = {abs/2503.12937},
  year         = {2025},
  url          = {https://doi.org/10.48550/arXiv.2503.12937},
  doi          = {10.48550/ARXIV.2503.12937},
  eprinttype    = {arXiv},
  eprint       = {2503.12937},
  bibsource    = {dblp computer science bibliography, https://dblp.org}
}

@article{sparl,
  author       = {Hanlin Wang and
                  Chak Tou Leong and
                  Jiashuo Wang and
                  Jian Wang and
                  Wenjie Li},
  title        = {{SPA-RL:} Reinforcing {LLM} Agents via Stepwise Progress Attribution},
  journal      = {CoRR},
  volume       = {abs/2505.20732},
  year         = {2025},
  url          = {https://doi.org/10.48550/arXiv.2505.20732},
  doi          = {10.48550/ARXIV.2505.20732},
  eprinttype    = {arXiv},
  eprint       = {2505.20732},
  bibsource    = {dblp computer science bibliography, https://dblp.org}
}

@article{gigpo,
  author       = {Lang Feng and
                  Zhenghai Xue and
                  Tingcong Liu and
                  Bo An},
  title        = {Group-in-Group Policy Optimization for {LLM} Agent Training},
  journal      = {CoRR},
  volume       = {abs/2505.10978},
  year         = {2025},
  url          = {https://doi.org/10.48550/arXiv.2505.10978},
  doi          = {10.48550/ARXIV.2505.10978},
  eprinttype    = {arXiv},
  eprint       = {2505.10978},
  bibsource    = {dblp computer science bibliography, https://dblp.org}
}

@article{treegrpo,
  author       = {Yuxiang Ji and
                  Ziyu Ma and
                  Yong Wang and
                  Guanhua Chen and
                  Xiangxiang Chu and
                  Liaoni Wu},
  title        = {Tree Search for {LLM} Agent Reinforcement Learning},
  journal      = {CoRR},
  volume       = {abs/2509.21240},
  year         = {2025},
  url          = {https://doi.org/10.48550/arXiv.2509.21240},
  doi          = {10.48550/ARXIV.2509.21240},
  eprinttype    = {arXiv},
  eprint       = {2509.21240},
  bibsource    = {dblp computer science bibliography, https://dblp.org}
}

@article{mtppo,
  author       = {Siliang Zeng and
                  Quan Wei and
                  William Brown and
                  Oana Frunza and
                  Yuriy Nevmyvaka and
                  Mingyi Hong},
  title        = {Reinforcing Multi-Turn Reasoning in {LLM} Agents via Turn-Level Credit
                  Assignment},
  journal      = {CoRR},
  volume       = {abs/2505.11821},
  year         = {2025},
  url          = {https://doi.org/10.48550/arXiv.2505.11821},
  doi          = {10.48550/ARXIV.2505.11821},
  eprinttype    = {arXiv},
  eprint       = {2505.11821},
  bibsource    = {dblp computer science bibliography, https://dblp.org}
}

@article{gtpo,
  title={Empowering Multi-Turn Tool-Integrated Reasoning with Group Turn Policy Optimization},
  author={Ding, Yifeng and Le, Hung and Han, Songyang and Ruan, Kangrui and Jin, Zhenghui and Kumar, Varun and Wang, Zijian and Deoras, Anoop},
  journal={arXiv preprint arXiv:2511.14846},
  year={2025}
}

@inproceedings{parallelbayes,
  author       = {Hanbin Hu and
                  Peng Li and
                  Jianhua Z. Huang},
  editor       = {Iris Bahar},
  title        = {Parallelizable Bayesian optimization for analog and mixed-signal rare
                  failure detection with high coverage},
  booktitle    = {Proceedings of the International Conference on Computer-Aided Design,
                  {ICCAD} 2018, San Diego, CA, USA, November 05-08, 2018},
  pages        = {98},
  publisher    = {{ACM}},
  year         = {2018},
  url          = {https://doi.org/10.1145/3240765.3240835},
  doi          = {10.1145/3240765.3240835},
  bibsource    = {dblp computer science bibliography, https://dblp.org}
}

@inproceedings{batchbayes,
  title={Batch Bayesian optimization via multi-objective acquisition ensemble for automated analog circuit design},
  author={Lyu, Wenlong and Yang, Fan and Yan, Changhao and Zhou, Dian and Zeng, Xuan},
  booktitle={International conference on machine learning},
  pages={3306--3314},
  year={2018},
  organization={PMLR}
}

@inproceedings{adollm,
  author       = {Yuxuan Yin and
                  Yu Wang and
                  Boxun Xu and
                  Peng Li},
  editor       = {Jinjun Xiong and
                  Robert Wille},
  title        = {{ADO-LLM:} Analog Design Bayesian Optimization with In-Context Learning
                  of Large Language Models},
  booktitle    = {Proceedings of the 43rd {IEEE/ACM} International Conference on Computer-Aided
                  Design, {ICCAD} 2024, Newark Liberty International Airport Marriott,
                  NJ, USA, October 27-31, 2024},
  pages        = {81:1--81:9},
  publisher    = {{ACM}},
  year         = {2024},
  url          = {https://doi.org/10.1145/3676536.3676816},
  doi          = {10.1145/3676536.3676816},
  bibsource    = {dblp computer science bibliography, https://dblp.org}
}

@article{anaflow,
  author       = {Mohsen Ahmadzadeh and
                  Kaichang Chen and
                  Georges G. E. Gielen},
  title        = {AnaFlow: Agentic LLM-based Workflow for Reasoning-Driven Explainable
                  and Sample-Efficient Analog Circuit Sizing},
  journal      = {CoRR},
  volume       = {abs/2511.03697},
  year         = {2025},
  url          = {https://doi.org/10.48550/arXiv.2511.03697},
  doi          = {10.48550/ARXIV.2511.03697},
  eprinttype    = {arXiv},
  eprint       = {2511.03697},
  bibsource    = {dblp computer science bibliography, https://dblp.org}
}

@article{eesizer,
  author       = {Chang Liu and
                  Danial Chitnis},
  title        = {EEsizer: LLM-Based {AI} Agent for Sizing of Analog and Mixed Signal
                  Circuit},
  journal      = {CoRR},
  volume       = {abs/2509.25510},
  year         = {2025},
  url          = {https://doi.org/10.48550/arXiv.2509.25510},
  doi          = {10.48550/ARXIV.2509.25510},
  eprinttype    = {arXiv},
  eprint       = {2509.25510},
  bibsource    = {dblp computer science bibliography, https://dblp.org}
}

@inproceedings{Artisan,
  author       = {Zihao Chen and
                  Jiangli Huang and
                  Yiting Liu and
                  Fan Yang and
                  Li Shang and
                  Dian Zhou and
                  Xuan Zeng},
  editor       = {Vivek De},
  title        = {Artisan: Automated Operational Amplifier Design via Domain-specific
                  Large Language Model},
  booktitle    = {Proceedings of the 61st {ACM/IEEE} Design Automation Conference, {DAC}
                  2024, San Francisco, CA, USA, June 23-27, 2024},
  pages        = {39:1--39:6},
  publisher    = {{ACM}},
  year         = {2024},
  url          = {https://doi.org/10.1145/3649329.3655903},
  doi          = {10.1145/3649329.3655903},
  bibsource    = {dblp computer science bibliography, https://dblp.org}
}

@article{easysize,
  author       = {Xinyue Wu and
                  Fan Hu and
                  Jani Babu Shaik and
                  Yi Zhao and
                  Xinfei Guo},
  title        = {EasySize: Elastic Analog Circuit Sizing via LLM-Guided Heuristic Search},
  journal      = {CoRR},
  volume       = {abs/2508.05113},
  year         = {2025},
  url          = {https://doi.org/10.48550/arXiv.2508.05113},
  doi          = {10.48550/ARXIV.2508.05113},
  eprinttype    = {arXiv},
  eprint       = {2508.05113},
  bibsource    = {dblp computer science bibliography, https://dblp.org}
}

@article{eraofexperience,
  title={Welcome to the era of experience},
  author={Silver, David and Sutton, Richard S},
  journal={Google AI},
  volume={1},
  year={2025}
}

@inproceedings{go,
title={Mixing Expert Knowledge: Bring Human Thoughts Back To the Game of Go},
author={Yichuan Ma and Linyang Li and Yongkang Chen and Peiji Li and Jiasheng Ye and Qipeng Guo and Dahua Lin and Kai Chen},
booktitle={The Thirty-ninth Annual Conference on Neural Information Processing Systems},
year={2025},
url={https://openreview.net/forum?id=baNBqpzvMT}
}

@misc{sketch,
      title={SketchThinker-R1: Towards Efficient Sketch-Style Reasoning in Large Multimodal Models}, 
      author={Ruiyang Zhang and Dongzhan Zhou and Zhedong Zheng},
      year={2026},
      eprint={2601.02825},
      archivePrefix={arXiv},
      primaryClass={cs.CV},
      url={https://arxiv.org/abs/2601.02825}, 
}

@article{verl,
  title   = {HybridFlow: A Flexible and Efficient RLHF Framework},
  author  = {Guangming Sheng and Chi Zhang and Zilingfeng Ye and Xibin Wu and Wang Zhang and Ru Zhang and Yanghua Peng and Haibin Lin and Chuan Wu},
  year    = {2024},
  journal = {arXiv preprint arXiv: 2409.19256}
}

@Misc{bayespython,
    author = {Fernando Nogueira},
    title = {{Bayesian Optimization}: Open source constrained global optimization tool for {Python}},
    year = {2014--},
    url = " https://github.com/bayesian-optimization/BayesianOptimization"
}

@misc{optbench,
      title={OPT-BENCH: Evaluating LLM Agent on Large-Scale Search Spaces Optimization Problems}, 
      author={Xiaozhe Li and Jixuan Chen and Xinyu Fang and Shengyuan Ding and Haodong Duan and Qingwen Liu and Kai Chen},
      year={2025},
      eprint={2506.10764},
      archivePrefix={arXiv},
      primaryClass={cs.AI},
      url={https://arxiv.org/abs/2506.10764}, 
}

@misc{npengine,
      title={NP-Engine: Empowering Optimization Reasoning in Large Language Models with Verifiable Synthetic NP Problems}, 
      author={Xiaozhe Li and Xinyu Fang and Shengyuan Ding and Linyang Li and Haodong Duan and Qingwen Liu and Kai Chen},
      year={2025},
      eprint={2510.16476},
      archivePrefix={arXiv},
      primaryClass={cs.AI},
      url={https://arxiv.org/abs/2510.16476}, 
}

@misc{openevolve,
  title = {{OpenEvolve}: An Open-Source Evolutionary Coding Agent},
  author = {Asankhaya Sharma},
  year = {2025},
  howpublished = {\url{https://github.com/algorithmicsuperintelligence/openevolve}},
  note = {Accessed: January 15, 2026}
}

@article{alphaevolve,
  title={AlphaEvolve: A coding agent for scientific and algorithmic discovery},
  author={Novikov, Alexander and V{\~u}, Ng{\^a}n and Eisenberger, Marvin and Dupont, Emilien and Huang, Po-Sen and Wagner, Adam Zsolt and Shirobokov, Sergey and Kozlovskii, Borislav and Ruiz, Francisco JR and Mehrabian, Abbas and others},
  journal={arXiv preprint arXiv:2506.13131},
  year={2025}
}

@article{kevin,
  title={Kevin: Multi-turn rl for generating cuda kernels},
  author={Baronio, Carlo and Marsella, Pietro and Pan, Ben and Guo, Simon and Alberti, Silas},
  journal={arXiv preprint arXiv:2507.11948},
  year={2025}
}
\end{multicols}


\clearpage
\appendix
\section{Details of ACS Circuits}
\label{apd:acs}
Our dataset comprises \textbf{12} practical analog circuit sizing tasks, implemented across multiple industrial process technologies and exhibiting significant diversity in both netlist complexity and problem scale (see Table~\ref{tab:netlist_tokens} for details of each circuit).

\begin{itemize}
    \item \textbf{ACCIA:} AC-coupled instrumentation amplifier (0.18-\textmu m process).
    \item \textbf{OPAMP1--4:} Four operational amplifiers utilizing different process nodes: TSMC 40nm Mixed-Signal RF, TSMC 0.18\textmu m RF, TSMC 0.18\textmu m Mixed-Signal, and SMIC 40nm Logic.
    \item \textbf{OTA:} Telescopic-cascode operational transconductance amplifier (TSMC 0.18\textmu m RF).
    \item \textbf{OTA1:} Cascode-compensated OTA from a pipeline ADC design, employing an opamp-sharing scheme (SMIC 0.18\textmu m).
    \item \textbf{ADC1--ADCL:} A family of analog-to-digital converters with 8--10 bit resolution, operating at sampling frequencies from 20 MHz to 1 GHz. Process technologies include SMIC 0.18\textmu m, TSMC 28nm, and TSMC 65nm.
    \item \textbf{SimLDO:} A low-dropout linear voltage regulator (LDO). This power-management circuit represents a key category of analog design focused on voltage regulation.
\end{itemize}

\begin{table}[ht]
\centering
\caption{Token lengths and numbers of tunable parameters for each circuit in our dataset. The tilde ($\sim$) denotes approximate values. OPAMP1--4 share similar netlist structures (with PDK variations), resulting in comparable token lengths and identical parameter counts.}
\label{tab:netlist_tokens}
\begin{tabular}{lccccccccc}
\toprule
\textbf{Circuit} & \textbf{ACCIA} & \textbf{ADC1} & \textbf{ADC2} & \textbf{ADC3} & \textbf{ADC4} & \textbf{ADCL} & \textbf{OTA} & \textbf{SimLDO} & \textbf{OPAMP1--4} \\
\midrule
Token Length & 12,208 & 5,244 & 4,054 & 4,735 & 5,418 & 3,550 & 1,185 & 3,289 & $\sim$800 \\
Tunable Parameters & 38 & 28 & 24 & 24 & 35 & 17 & 10 & 37 & 19 \\
\bottomrule
\end{tabular}
\end{table}

\section{Implementation Details of Unified Turn-level Reward Function}
\label{apd:reward_details}

The normalized score function $p_j(v_j, s_j)$ for each specification type is defined as follows. Let $\tau_j^{\ell}$ and $\tau_j^{u}$ be tolerance thresholds that define smooth transitions between acceptable and unacceptable performance. These thresholds are set proportionally to the specification values (e.g., $\tau_j^{\ell} = \alpha |s_j|, \tau_j^{u} = \beta |s_j|$) to maintain physical meaningfulness.


\textbf{1. Lower-bound specification} ($v_j$ should satisfy $v_j \geq s_j$):
\begin{align}
p_j^{\text{lower}}(v_j, s_j) &= 
\begin{cases} 
0, & v_j < s_j - \tau_j^{\ell} \\
\big(\frac{v_j - (s_j - \tau_j^{\ell})}{\tau_j^{\ell}}\big)^2, & s_j - \tau_j^{\ell} \leq v_j < s_j \\
1, & v_j \geq s_j
\end{cases} 
\end{align}

\textbf{2. Upper-bound specification} ($v_j$ should satisfy $v_j \leq s_j$):
\begin{align}
p_j^{\text{upper}}(v_j, s_j) &= 
\begin{cases} 
1, & v_j \leq s_j \\ 
\big(\frac{s_j + \tau_j^{u} - v_j}{\tau_j^{u}}\big)^3, & s_j < v_j \leq s_j + \tau_j^{u} \\
0, & v_j > s_j + \tau_j^{u}
\end{cases} 
\end{align}

\textbf{3. Range specification} ($v_j$ should fall within an interval $s_j = [\ell_j, u_j]$):
\begin{align}
p_j^{\text{range}}(v_j, s_j) &=
\begin{cases} 
0, & v_j < \ell_j - \tau_j^{\ell} \\
\big(\frac{v_j - (\ell_j - \tau_j^{\ell})}{\tau_j^{\ell}}\big)^2, & \ell_j - \tau_j^{\ell} \leq v_j < \ell_j \\
1, & \ell_j \leq v_j \leq u_j \\
\big(\frac{u_j + \tau_j^{u} - v_j}{\tau_j^{u}}\big)^3, & u_j < v_j \leq u_j + \tau_j^{u} \\
0, & v_j > u_j + \tau_j^{u}
\end{cases} 
\end{align}

These continuous, piecewise-defined functions—composed of constant, quadratic, and cubic segments—are constructed to remain within $[0, 1]$, ensuring that their geometric mean (Eq.~\ref{eq:performance_reward}) also falls within $[0, 1]$ and provides a normalized overall score. This design delivers smooth penalties near specification boundaries, encouraging the LLM agent to robustly meet targets rather than marginally violating them.

\section{Training Setup Details}
\label{apd:trainingsetup}

We provide a comprehensive summary of the hyperparameters and implementation details used for all RL training paradigms. All experiments are conducted using the Qwen3-30B-A3B-Instruct model as the base policy, and we conducted our experiments based on VeRL codebase \cite{verl}. We use a global prompt batch size of 32 and learning rate of $1\times10^{-6}$ for 1 epoch. For multi-turn paradigms (TL-GRPO and Trajectory-Level GRPO), each query is allowed a maximum of $T=5$ tool-call turns. During the rollout phase, the sampling group size $G$ is 8 (applied at turn level for TL-GRPO and at sequence level for Trajectory-Level GRPO and Single-Turn GRPO) with a temperature of 1.0 and top-$p$ sampling of 0.95. To accommodate long reasoning traces and tool outputs, we set the maximum prompt length to 15,000 tokens (for long circuit netlist content) and the maximum response length per turn to 4,096 tokens. Following practices from \cite{dapo}, we employ  clip ratios $\varepsilon_{\text{low}}=0.2$ and $\varepsilon_{\text{high}}=0.28$, and we do not apply KL-divergence regularization. 

\section{Analysis of Optimization Dynamics Across Turns}
\label{apd:turn_analysis}
\begin{figure*}[ht]
    \centering
    \vspace{5pt}
    \includegraphics[width=\linewidth]{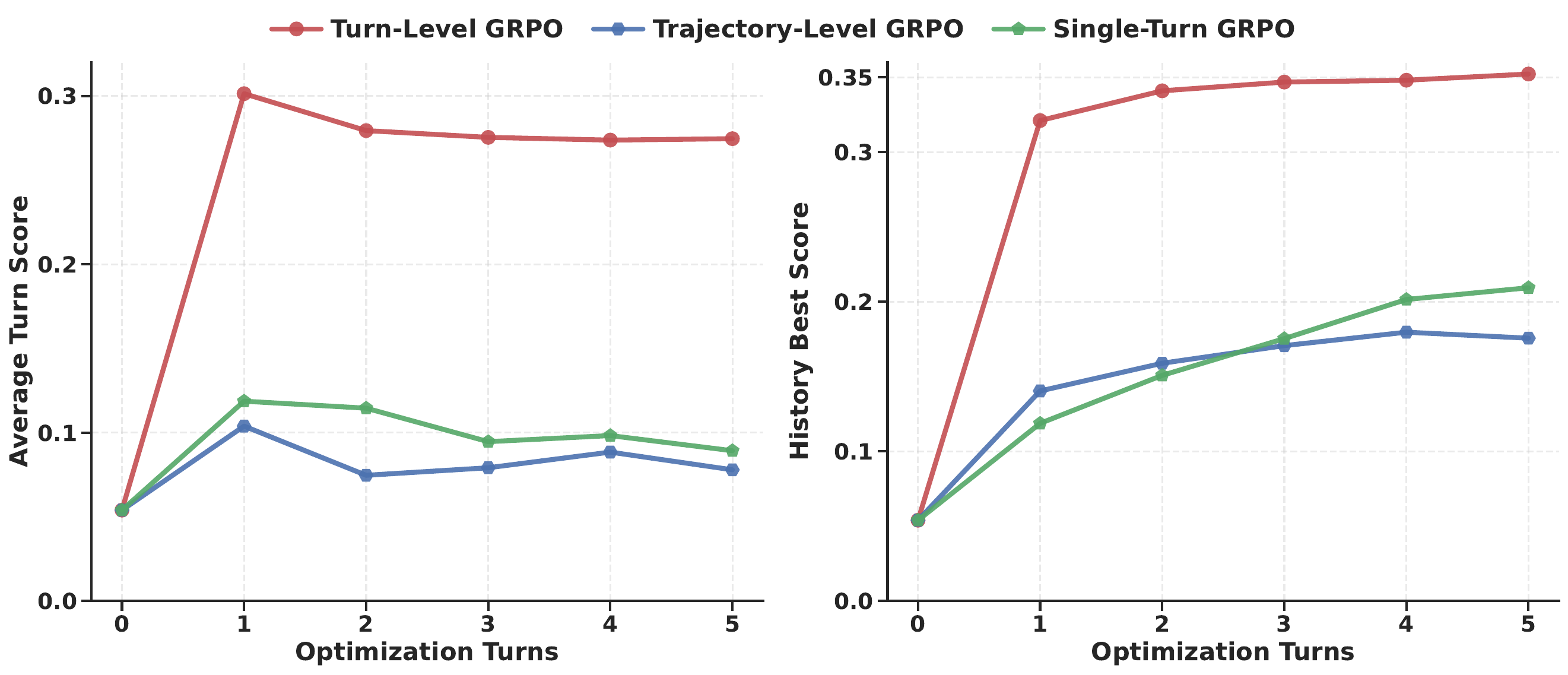}
\caption{Comparison of turn-level performance for models trained with three RL algorithms on out-of-domain tasks. The turn-0 score represents the performance of initial values. Each line shows the average across all evaluation data. }

    \label{fig:turn_analysis}
\vspace{5pt}  
\end{figure*}

We analyze the evaluation behavior of models trained with the three RL algorithms to understand how they utilize the available optimization turns. \Cref{fig:turn_analysis} presents two key metrics measured on out-of-domain tasks: (1) the average score achieved at each turn, and (2) the history-best score (i.e., the maximum reward observed up to that turn).

Three key observations emerge from this analysis:

\textbf{Effectiveness of Initial Optimization:} All three methods show a substantial increase in average turn score from turn 0 (initial random parameters) to turn 1, demonstrating that even a single optimization step guided by LLM reasoning significantly improves design quality. This immediate improvement distinguishes reasoning-guided iterative optimization from black-box methods, which typically require multiple exploratory samples before achieving comparable gains.

\textbf{Exploration in Later Turns:} After the first turn, average turn scores plateau or exhibit minor fluctuations, indicating that subsequent turns are primarily used for local exploration and refinement rather than consistently achieving higher scores. This pattern reflects a realistic balance between exploitation of current knowledge and exploration of the design space.

\textbf{Progressive Improvement in Best Design:} Despite the plateau in average turn scores, the history-best score consistently improves across turns for all methods, confirming that additional optimization attempts increase the probability of discovering better designs. Notably, TL-GRPO achieves both higher average turn scores and faster accumulation of best scores throughout the optimization process.

These observations validate that TL-GRPO's turn-level optimization strategy effectively guides the agent to make higher-quality proposals at each step, leading to more efficient discovery of optimal designs within the fixed turn budget.

\section{Data Examples}
\label{apd:dataexample}

We provide an example of our synthetic training data (\Cref{fig:example_input}), the tool schema used for simulation (\Cref{fig:example_schema}), and a sequence of five turns from our 30B model trained with TL-GRPO algorithm (\cref{fig:example_response_turn_1,fig:example_response_turn_2,fig:example_response_turn_3,fig:example_response_turn_4,fig:example_response_turn_5}). These examples illustrate the reasoning-guided iterative optimization process for analog circuit sizing.

\begin{figure}[H]
    \centering
    \begin{tcolorbox}[
        enhanced,
        colframe=deepblue,
        colback=lightblue!10,
        boxrule=1pt,
        arc=2mm,
        title={Example Schema of Analog Circuit Simulation Tool},
        fonttitle=\bfseries,
        top=0pt,
        bottom=0pt,
        left=0pt,
        right=0pt
    ]
    \begin{lstlisting}[basicstyle=\ttfamily\tiny] 
tool_schema:
  type: "function"
  function:
    name: "acs_simulate"
    description: |
      Simulates an analog circuit using Cadence Spectre.
      The function accepts a dictionary of transistor parameters. Depending on the specific circuit topology,these may include widths (e.g., "w1", "w2", ...), lengths (e.g., "l1", "l2", ...), custom-named sizing parameters (e.g., WBOOTPM1", ...), or passive component parameters such as capacitors and resistors. The exact set and naming of available parameters are determined by the circuit netlist. Each parameter value must be a string with appropriate unit. Please use "u" for sizing parameters, "k"(kilo-ohm) for resistors, and "p"(pF) for capacitors. For example, use "2u" for 2 micrometers, "10k" for 10 kilo-ohms, and "5p" for 5 picofarads.
      The simulation runs in an isolated temporary directory. Upon completion, it may extract various performance metrics, which could include but are not limited to:
      **DC/Bias Metrics:**
      - vout: Output voltage in Volts

      **Frequency Response:**
      - gain: DC small-signal gain in dB
      - gbw: unity-gain bandwidth in Hz
      - pm: phase margin in degree
      - gm: Gain margin in dB
      - PhaseMarginFrequency: Frequency at phase margin in Mhz
      - GainMarginFrequency: Frequency at gain margin in Mhz

      **Power Supply Rejection (PSR):**          
      - PSR@specific_freq: PSR at a specific frequency in dB(e.g., PSR@10Hz, PSR@10kHz
      - PSR_peak: Peak PSR in dB

      **Noise Performance:**
      - irn: input reference noise in Voltage
      - OutNoise@specific_freq: Output noise at a specific frequency in Voltage/sqrt(Hz)

      **Power Consumption:**
      - pw: total power consumption in Watts

      **Data Converter Metrics (if applicable):**
      - enob: effective number of bits (bits)
      - sfdr: spurious-free dynamic range (dBc)
      - snr: signal-to-noise ratio (dB)
      Returns a dictionary with the available simulation metrics as strings. The specific metrics returned depend on the circuit type and simulation setup.

    parameters:
      properties:
        variables:
          type: "object"
          description: |
            A dictionary mapping transistor parameter names to their values as strings with units. The available parameters depend on the circuit topology and may include: Generic width/length names (e.g., "w1", "w2", ..., "l1", "l2", ...), Custom netlist-defined names (e.g., "WCOMPN0", "WBOOTPM1", "WDACNM2", "FinWidth_Nbias0", etc.), Passive component parameters (e.g., Cap_z3, Res_bias1) if applicable.  
            Example with generic names: {"w1": "2u", "w2": "3u", ..., "wN": "4u"}
            The specific parameter names and their count are determined by the circuit netlist. All values must be strings with appropriate units, not numbers.
    \end{lstlisting}
    \end{tcolorbox}
    \caption{Example tool schema for analog circuit simulation.}
    \label{fig:example_schema}
\end{figure}

\begin{figure}[htbp]
    \centering
    \begin{tcolorbox}[
        enhanced,
        colframe=deepblue,
        colback=lightblue!10,
        boxrule=1pt,
        arc=2mm,
        title={Example Synthetic Input Query with Initial Points (score = 0)},
        fonttitle=\bfseries,
        top=0pt,
        bottom=0pt,
        left=0pt,
        right=0pt
    ]
    \begin{lstlisting}[basicstyle=\ttfamily\scriptsize] 
You are a senior analog integrated circuit design engineer responsible for performing a transistor sizing optimization (Analog Circuit Sizing) task.
## Task Description
Your goal is to optimize the width or length parameters of all specified MOSFET transistors in a given circuit netlist through iterative simulation and parameter adjustment, so that they meet a series of preset performance metric requirements.
## Circuit Netlist
The following is the SPICE netlist description of the circuit, which defines the circuit topology, component connections, and the parameters you need to optimize:
```spice
// This circuit should be designed and optimized based on the TSMC 40nm Mixed Signal RF SALICIDE process library. In the circuit netlist, `nch` represents the NMOS transistor under this process with a DC bias voltage of 1.8V, and `pch` represents the PMOS transistor under this process with a DC bias voltage of 1.8V.
subckt opamp net_vb_n net_vb_p net_vin net_vip net_vo vdd vss
    I0_M5 (net_vb_n net_vb_n vss vss) nch l=120n w=w1 m=1 nf=4
    I0_M4 (net4 net_vb_n vss vss) nch l=120n w=w2 m=1 nf=4
    I0_M1 (net5 net_vin net4 vss) nch l=120n w=w3 m=1 nf=4
    I0_M0 (net_1 net_vip net4 vss) nch l=120n w=w4 m=1 nf=4
    I0_M3 (net5 net5 vdd vdd) pch l=120n w=w5 m=1 nf=4
    I0_M2 (net_1 net5 vdd vdd) pch l=120n w=w6 m=1 nf=4
    I1_M4 (net_vb_p net_vb_p vdd vdd) pch l=120n w=w7 m=1 nf=4
    I1_M1 (net_2 net70 vss vss) nch l=120n w=w8 m=1 nf=4
    I1_M0 (net70 net70 vss vss) nch l=120n w=w9 m=1 nf=4
    I1_M3 (net_2 net_vb_p vdd vdd) pch l=120n w=w10 m=1 nf=4
    I1_M2 (net70 net_1 vdd vdd) pch l=120n w=w11 m=1 nf=4
    I2_M2 (net_vb_n net_vb_n vss vss) nch l=120n w=w12 m=1 nf=4
    I2_M0 (net_vo net_vb_n vss vss) nch l=120n w=w13 m=1 nf=4
    I2_M1 (net_vo net_2 vdd vdd) pch l=120n w=w14 m=1 nf=4
    I3_M4 (net_vb_p net_vb_p vdd vdd) pch l=120n w=w15 m=1 nf=4
    I3_M1 (net_vo net71 vss vss) nch l=120n w=w16 m=1 nf=4
    I3_M0 (net71 net71 vss vss) nch l=120n w=w17 m=1 nf=4
    I3_M3 (net_vo net_vb_p vdd vdd) pch l=120n w=w18 m=1 nf=4
    I3_M2 (net71 net_1 vdd vdd) pch l=120n w=w19 m=1 nf=4
ends opamp
```
## Parameters to Optimize
The parameters to be optimized for this circuit are all width parameters represented by w1, w2, ... w19. The value range for each parameter is from 0.9 μm to 9 μm.

## Initial Parameter Values
w1: 5.9 μm w2: 5.99 μm w3: 1.36 μm w4: 1.43 μm w5: 5.25 μm w6: 7.03 μm w7: 8.4 μm w8: 6.54 μm w9: 2.99 μm w10: 1.2 μm w11: 7.11 μm w12: 5.42 μm w13: 3.17 μm w14: 5.94 μm w15: 5.07 μm w16: 2.05 μm w17: 1.5 μm w18: 2.77 μm w19: 7.48 μm

## Performance Targets
The optimized circuit must meet the following performance metrics:
- Gain > 79.14 dB
- Gain-Bandwidth Product (GBW) > 0.59 MHz
- Total Power Consumption (Power) < 17.77 μW
- Phase Margin (PM) > 70.95 degrees
// ... (details about task format and specific requirements are omitted for brevity)
// Initial value with simulation result
Arguments: {"variables": {"w1": "5.9u", "w2": "5.99u", "w3": "1.36u", "w4": "1.43u", "w5": "5.25u", "w6": "7.03u", "w7": "8.4u", "w8": "6.54u", "w9": "2.99u", "w10": "1.2u", "w11": "7.11u", "w12": "5.42u", "w13": "3.17u", "w14": "5.94u", "w15": "5.07u", "w16": "2.05u", "w17": "1.5u", "w18": "2.77u", "w19": "7.48u"}}
Simulation Result:
{'gain': '64.5553', 'gbw': '23314.7', 'pw': '6.29918e-06', 'pm': '89.5388'}
    \end{lstlisting}
    \end{tcolorbox}
    \caption{An example of our synthetic query with initial points.}
    \label{fig:example_input}
\end{figure}

\begin{figure}[htbp]
    \centering
    \begin{tcolorbox}[
        enhanced,
        colframe=deepblue,
        colback=lightblue!10,
        boxrule=1pt,
        arc=2mm,
        title={Example LLM Response Turn 1 (score=0.24)},
        fonttitle=\bfseries,
        top=0pt,
        bottom=0pt,
        left=0pt,
        right=0pt
    ]
    \begin{lstlisting}[basicstyle=\ttfamily\scriptsize] 
<Analyze>
Based on the simulation results with the initial parameters, the current circuit performance does not meet the design targets. The specific analysis is as follows:

1.  **Gain**: The current gain is 64.5553 dB, far below the target of 79.14 dB. The gain is primarily determined by the ratio of the transconductance (gm) to the output impedance (ro) of the differential pairs (e.g., M1-M2 and M0-M3). Currently, their forward current (controlled by w3, w4, w5, w6, etc.) is insufficient to generate adequate transconductance. Furthermore, their widths (w3, w4) are set too small, leading to the overall low gain.

2.  **Gain-Bandwidth Product (GBW)**: The current GBW is 23.315 kHz, far from the target of 0.59 MHz. GBW is mainly determined by the dominant pole (defined by the input stage's input capacitance and transconductance) and the second pole. The small sizes of the input stage transistors (e.g., M1-M2, M0-M3, etc.) result in low transconductance, and the excessive load capacitance (determined by the equivalent capacitive structure) leads to an extremely low dominant pole frequency.

3.  **Power Consumption**: The current power consumption is 6.299 μW, which is lower than the target of 17.77 μW, leaving significant margin. However, this is partly because the sizes of many key transistors (e.g., M3-M4, M5-M6, etc.) are set too large, while some other key transistors (e.g., M1-M2) are too small, leading to inefficiency. Although the total power is low, it still needs optimization to meet the low-power requirement while ensuring other specifications.

4.  **Phase Margin**: The current phase margin is 89.54°, higher than the target of 70.95°. This high margin, however, is caused by excessively high impedance and excessively small capacitance (or small sizes). Consequently, the dominant pole frequency is too low, resulting in minimal circuit bandwidth. Additionally, the uneven distribution of small-signal "branch" currents leads to a mismatch between open-loop gain and bandwidth, causing overall instability.

5.  **Sizes of key transistors (e.g., w1, w2, w3, w4, w5, w6)**: These are the core of the differential pair. Their values (w1~w6) directly relate to the base current (of M0~M5). Currently, they are set unevenly and do not reach the ideal proportions. Particularly, the sizes of w3 and w4 (input transistors) are too small, leading directly to insufficient gm, which is the primary cause of low gain.

6.  **Current mirrors and tail current transistors (e.g., w1~w2, w7~w10)**: Their sizes (w1~w6) set the reference current, but their matching is poor currently, and the total current is not optimized. Meanwhile, the sizes of source follower transistors (w12~w13, w14~w15) also need adjustment to improve tail current stability and increase output impedance.

7.  **Transistors (w8~w11)**: These are auxiliary transistors for bias generation or current mirroring. Their sizes (e.g., w8~w11) are set too high in the original values, causing wasted power, yet they play a crucial role in the loop.

The current layout and parameter values are severely imbalanced. Manual adjustments must focus on **increasing the transconductance of key transistors**, **optimizing current distribution**, and **balancing the sizes of various branches**, while also **re-planning the combination of "current," "complementary pairs," and "tail current."**

In summary, the current design is far from the target. Systematic optimization is required, starting with correcting the differential pair sizes (w3, w4, w5, w6), adjusting the bias (w1, w2, w7, w8), optimizing current mirrors and mirror ratios (w9~w10), and transistors related to the dominant pole (w13~w17).

The next round of adjustments will focus on improving gain and bandwidth while controlling power consumption.
</Analyze>
<tool_call>
Function: acs_simulate
Arguments: {"variables": {"w1": "2.7u", "w2": "2.7u", "w3": "6.5u", "w4": "6.5u", "w5": "5.8u", "w6": "5.8u", "w7": "1.8u", "w8": "1.7u", "w9": "0.9u", "w10": "1.8u", "w11": "1.8u", "w12": "1.0u", "w13": "3.3u", "w14": "3.3u", "w15": "1.7u", "w16": "1.7u", "w17": "0.9u", "w18": "1.8u", "w19": "1.8u"}}
</tool_call>
<tool_response>
Simulation Result:
{'gain': '68.4952', 'gbw': '93849.3', 'pw': '9.54376e-06', 'pm': '89.5635'}
</tool_response>
    \end{lstlisting}
    \end{tcolorbox}
    \caption{An example of LLM response: Turn 1.}
    \label{fig:example_response_turn_1}
\end{figure}

\begin{figure}[htbp]
    \centering
    \begin{tcolorbox}[
        enhanced,
        colframe=deepblue,
        colback=lightblue!10,
        boxrule=1pt,
        arc=2mm,
        title={Example LLM Response Turn 2 (score=0.42)},
        fonttitle=\bfseries,
        top=0pt,
        bottom=0pt,
        left=0pt,
        right=0pt
    ]
    \begin{lstlisting}[basicstyle=\ttfamily\tiny] 
<Analyze>
According to the simulation results after this parameter adjustment, the circuit performance has improved significantly compared to the initial values but still fails to meet the design targets, requiring further optimization. The detailed analysis is as follows:
1.  **Gain**: The current gain is 68.4952 dB, still below the target of 79.14 dB, with a gap of about 10.64 dB. It is primarily determined by the ratio of the transconductance (gm) to the output impedance (ro) of the differential pairs (M1-M2, M0-M3). Currently, w3 and w4 (input transistors) have been adjusted to 6.5u, providing a slight increase, but their widths are still insufficient. Moreover, their matching with w5 and w6 (PMOS) is suboptimal, leading to inadequate overall transconductance. Additionally, the sizes of current mirrors (e.g., w1~w2) and tail current transistors (w12~w13) are set too small to support high gain.

2.  **Gain-Bandwidth Product (GBW)**: The current GBW is 93.85 kHz, still far from the target of 0.59 MHz (590 kHz), with a gap of 496 kHz. It is mainly influenced by the dominant pole frequency (determined by the gate-source capacitance and transconductance of the input transistors). Although significantly improved from the initial value (23.3 kHz), the related transistors (e.g., w3~w4) are still relatively small, and the overall load capacitance and current configuration are not ideal. Furthermore, the small size of the output stage (e.g., M13~M14) also limits speed.

3.  **Power Consumption**: The current power consumption is 9.54 μW, which is close to but still within the target range (<17.77 μW), leaving a significant margin. This benefits from reducing the sizes of several "tail current" transistors (e.g., w1~w2, w7~w8, w12~w13) to around 1u-2u. However, the device current is not sufficiently amplified in critical paths (e.g., M3~M4, M5~M6). Therefore, based on this, the sizes of main current paths should be reasonably increased to improve gain and speed while ensuring total power does not exceed the limit.

4.  **Phase Margin**: The current phase margin is 89.56°, higher than the target of 70.95°, and is very robust. This indicates good stability, meaning there is sufficient margin between the dominant and second poles, and the current circuit shows no significant oscillation tendency. However, with the current parameter settings, this high margin comes "at the cost of bandwidth," as the bandwidth remains too low, representing a case of "high margin but low efficiency."

5.  **Analysis of Key Transistor Sizes**:
    *   **w1~w2 (tail current transistors)**: Set to 2.7u, but they are mainly for biasing. Their small values contribute insufficiently to increasing gain, and their matching with w3~w4 is not aligned.
    *   **w3~w4 (input transistors)**: Set to 6.5u, crucial for improving gain, but their size needs further amplification.
    *   **w5~w6 (PMOS gate current mirrors)**: Their size is 5.8u, and their cooperation with w3~w4 is key. Currently, their ratio (w5/w3 = 5.8/6.5 ≈ 0.89) is less than ideal, making it difficult to generate sufficient transconductance and needs improvement.
    *   **w7~w10 (bias transistors)**: Set to 1.7u~1.8u, but their current is small, mainly for biasing rather than the main path. They need reasonable matching to support all transistors.
    *   **w11~w12 (current mirrors)**: In the current settings, their sizes are 1.0u~1.0u, insufficient to drive adequate current, which is one of the main reasons for the suboptimal performance.
    *   **w13~w14 (output stage transistors)**: Set to 3.3u. Although they are amplifying transistors, their mismatch with w15~w16 and w17~w18 leads to insufficient transconductance.
    *   **w15~w17 (bias and tail current)**: Their sizes are set too small (all in the 0.9u~1.7u range), making it difficult to support the required current.
    *   **w16~w17 (common-mode feedback auxiliary)**: Currently set to only 1.7u and 0.9u, used for stabilizing the output level, but their insufficient size affects overall performance.

6.  **Optimization Direction Summary**:
    *   **Primary Objective**: Increase gain (+~10.6 dB) and GBW (+~496 kHz).
    *   **Main Approach**: Increase the sizes of w3, w4 (differential input pair) and w5, w6 (symmetric PMOS), while adjusting w1~w2 and w12~w13 (tail current) to coordinate the improvement.
    *   **Secondary Objective**: Continue optimizing the size ratios of key transistors, especially w13~w14, to enhance their drive capability for the output stage, without significantly increasing power consumption (currently 9.54 μW).
    *   **Third Objective**: Adjust the reasonable allocation of various current mirror ratios (e.g., w7~w10, w15~w17) to ensure they provide sufficient "pumping" current for critical paths, while avoiding setting their sizes too small and dragging down overall performance.

In summary, although the current performance is better than the initial state, substantial size optimization is still needed, especially strategic reconfiguration of the differential pair (w3~w6), current mirrors (w5~w6, w13~w14), and tail current structure (w1~w2, w12~w13). Through this round of adjustments, we can begin to approach the target while maintaining the robustness of the phase margin.
</Analyze>
<tool_call>
Function: acs_simulate
Arguments: {"variables": {"w1": "1.0u", "w2": "1.0u", "w3": "8.0u", "w4": "8.0u", "w5": "7.4u", "w6": "7.4u", "w7": "1.5u", "w8": "1.3u", "w9": "0.9u", "w10": "1.5u", "w11": "1.5u", "w12": "1.0u", "w13": "3.7u", "w14": "3.7u", "w15": "1.5u", "w16": "1.5u", "w17": "0.9u", "w18": "1.7u", "w19": "1.7u"}}
</tool_call>
<tool_response>
Simulation Result:
{'gain': '69.948', 'gbw': '166584', 'pw': '1.08715e-05', 'pm': '88.9833'}
    \end{lstlisting}
    \end{tcolorbox}
    \caption{An example of LLM response: Turn 2.}
    \label{fig:example_response_turn_2}
\end{figure}

\begin{figure}[htbp]
    \centering
    \begin{tcolorbox}[
        enhanced,
        colframe=deepblue,
        colback=lightblue!10,
        boxrule=1pt,
        arc=2mm,
        title={Example LLM Response Turn 3 (score=0.44)},
        fonttitle=\bfseries,
        top=0pt,
        bottom=0pt,
        left=0pt,
        right=0pt
    ]
    \begin{lstlisting}[basicstyle=\ttfamily\tiny] 
<Analyze>
According to the simulation results after this parameter adjustment, the circuit performance has improved compared to the previous round (the second round for `w1~w19`), but it **still has a significant gap** from the target performance and requires further optimization. The detailed analysis is as follows:

1.  **Gain**: The current gain is 69.948 dB, still far below the target of 79.14 dB, with a gap of 9.19 dB. This is the core issue in the current design. Gain improvement relies on the product of the transconductance (gm) and output impedance (ro) of the differential pair. Currently, the core transistors (w3, w4) are only 8.0u, still insufficient compared to the target. Although their size has increased from before (6.5u), it remains below the ideal value (around 9.0u or higher). Furthermore, the width of M5-M6 (PMOS) (7.4u) has not reached the optimal ratio, and their matching with M3-M4 (NMOS) still needs optimization, resulting in insufficient transconductance to support high gain.

2.  **Gain-Bandwidth Product (GBW)**: The current GBW is 166.58 kHz, showing a clear improvement from the last round (93.85 kHz) (+72.73 kHz), but it still has a large distance from the target of 0.59 MHz (590 kHz), with a gap of 423.42 kHz. It is primarily dominated by the dominant pole (determined by the input stage's input capacitance and transconductance). The current combination of the input pair (w3, w4) and tail current (w1, w2) sizes increases the gm, while the load capacitance (from M5, M6, M13, M14, etc.) also increases, but the overall circuit still cannot support high bandwidth. Additionally, the cascode structure (e.g., M13-M14 as current mirrors) has a small size (3.7u), resulting in insufficient drive capability, and its "compensation" potential is not fully realized. There may also be redundant capacitance that needs fine-tuning.

3.  **Power Consumption**: The current power consumption is 10.87 μW, still below the target of 17.77 μW, leaving a significant margin (available increase of about 6.9 μW). However, the allocation has increased compared to the last round (9.54 μW), mainly due to increasing the sizes of M3~M4 (6.5u), M5~M6 (5.8u), and M13~M14 (3.3u). However, the efficiency of the current core path (e.g., w3~w4) is still not optimal; approximately 1.5μA current with 8.0u transistors is not at its best. Meanwhile, auxiliary transistors (e.g., w1~w2, w8~w10) sizes remain low. While a small further increase is possible, it must be ensured that performance breakthroughs are achieved without further increasing power consumption (maximum allowed is 17.77 μW).

4.  **Phase Margin**: The current PM is 88.98°, very robust and far exceeding the target of 70.95°. This indicates good stability of the circuit under the current structure, with no signs of imminent oscillation. However, the contradiction between high margin and low GBW persists (i.e., "large phase margin but low frequency") due to the unbalanced impedance and capacitance of the current input stage. The "capacitance-current" relationship has not been optimized to its best.

5.  **Key Transistor Parameter Analysis**:
    *   **w3~w4 (NMOS input pair)**: Set to 8.0u, currently the "short board." Based on matching the current M5~M6 (7.4u), their transconductance (gm) has not reached the level required for high gain.
    *   **w5~w6 (PMOS)**: Currently 7.4u, not ideally matched with w3~w4 (8.0u). If their "current mirror" characteristics are not matched, it will cause gain degradation. Their width-to-length ratio determines their transconductance and current ratio.
    *   **w1~w2 (tail current)**: Set to 1.0u, a key reference current transistor. Its small value cannot drive sufficient current, though it is slightly better than the previous round, but must be increased.
    *   **w13~w14 (output stage transistors)**: Set to 3.7u, key transistors for driving the output. Their size affects both gain and bandwidth, but being too small (only 3.7u) limits performance.
    *   **w15~w17 (auxiliary bias transistors)**: Set to 1.5u~0.9u, mainly for stabilizing the bias source current, but their values contribute to the whole. Currently, being too small may affect critical paths.
    *   **w7~w10 (bias current mirrors)**: Their size is 1.5u~1.5u, small but "reasonable" for biasing, acceptable, but insufficient to support large current demands.
    *   **w11~w12**: Part of the key current mirror section, their size is 1.5u~1.0u, which might be one reason for "mirroring error."
    *   **w16~w17**: Set to 1.5u and 0.9u, small currents but must be maintained.
    *   **w18~w19**: Set to 1.7u~1.7u, serving as current mirrors or matching transistors, matched with w13~w14, but their size is too small.

6.  **Optimization Direction Summary**:
    *   **Most Urgent Task**: Take **gain** (+9.2 dB) and **GBW** (+423.4 kHz) as the core improvement targets. Reasonably enlarge w3, w4, w5, w6, w13, w14, and adjust their current matching without increasing power consumption (≤17.77 μW).
    *   **Core Path Optimization**: Further increase w3~w4 (expected to reach 8.5-9.0u) and w5~w6 (should be 7.5-8.0u) to match their current sources while improving the transconductance-current ratio.
    *   **Increase Key Transistor (w13~w14) Sizes**: As output stage driver transistors, their width needs to be increased simultaneously to enhance current, supporting higher gain and larger bandwidth.
    *   **Utilize Power Margin**: Current power is 10.87 μW, with about 6.9 μW incremental available. Under reasonable allocation, **the sizes of transistors like w3~w6, w13~w14 can be increased**, achieving a "low-cost" performance leap.
    *   **Pay Attention to Matching and Proportional Relationships**: Ensure the width-to-length ratios of w3~w4 and w5~w6 are matched, and their current ratios are coordinated with w1~w2. Simultaneously, ensure the matching of various current mirrors (e.g., w7~w10, w15~w17).
    *   **Avoid "Saturated" Adjustments**: Current parameters already seem "tight" (e.g., w1, w2=1.0u). Although adjustable, avoid blind increases.

In summary, although current parameters have improved compared to the first two rounds, there is still a significant distance from the target. **Systematically** and reasonably enlarging the sizes of core transistors (w3~w6) and driver transistors (w13~w14), while utilizing the remaining power margin, is expected to achieve performance targets in the next simulation through strategic adjustment combinations. More precise control is anticipated to achieve a "two-stage leap" in gain and bandwidth.
</Analyze>
<tool_call>
Function: acs_simulate
Arguments: {"variables": {"w1": "1.0u", "w2": "1.0u", "w3": "9.0u", "w4": "9.0u", "w5": "8.7u", "w6": "8.7u", "w7": "1.5u", "w8": "1.2u", "w9": "0.9u", "w10": "1.5u", "w11": "1.5u", "w12": "1.0u", "w13": "4.0u", "w14": "4.0u", "w15": "1.5u", "w16": "1.4u", "w17": "0.9u", "w18": "1.7u", "w19": "1.7u"}}
</tool_call>
<tool_response>
Simulation Result:
{'gain': '70.0745', 'gbw': '178029', 'pw': '1.11515e-05', 'pm': '88.8258'}
    \end{lstlisting}
    \end{tcolorbox}
    \caption{An example of LLM response: Turn 3.}
    \label{fig:example_response_turn_3}
\end{figure}

\begin{figure}[htbp]
    \centering
    \begin{tcolorbox}[
        enhanced,
        colframe=deepblue,
        colback=lightblue!10,
        boxrule=1pt,
        arc=2mm,
        title={Example LLM Response Turn 4 (score=0.48)},
        fonttitle=\bfseries,
        top=0pt,
        bottom=0pt,
        left=0pt,
        right=0pt
    ]
    \begin{lstlisting}[basicstyle=\ttfamily\tiny] 
<Analyze>
According to the simulation results after this parameter adjustment, the circuit performance has again achieved a small but visible improvement on the existing basis, but it is **still far from the target** and requires more precise optimization. The detailed analysis is as follows:
1.  **Gain**: The current gain is 70.0745 dB, only 0.1265 dB higher than the last round (69.948 dB), still 9.0655 dB short of the target 79.14 dB, making it the primary current issue. The improvement is very slow and almost negligible. The main reasons are: although w3~w4 (8.0u → 9.0u) have been increased, the change from 8.0u to 9.0u brings limited transconductance improvement; and w5~w6 have also increased from 7.4u to 8.7u, but still insufficient to form a high enough transconductance-impedance ratio with w3~w4 (9.0u). Moreover, with current parameters, the relationship between channel length (fixed at 120n) and transconductance (gm) is nonlinear, showing significant diminishing marginal effects.

2.  **Gain-Bandwidth Product (GBW)**: The current GBW is 178.03 kHz, slightly higher than the last round (166.58 kHz) (+11.45 kHz), but still has a huge gap of 411.97 kHz compared to the target 0.59 MHz (590 kHz). Although improved due to increased sizes of w3~w4 and w5~w6, it still cannot support high bandwidth, mainly constrained by the following points:
    *   The transconductance (gm) of the input stage transistors has improved, but its ratio to the input capacitance (determined by gate and substrate capacitance) is still not ideal.
    *   The loading effect of "high impedance" nodes (e.g., M13~M14) has not been effectively offset; their "load" drags down frequency response.
    *   The size of current mirrors (e.g., w13~w14) is only 4.0u, their output impedance is limited, and their input-output matching is insufficient.
    *   The current design does not introduce "Miller capacitance" or "compensation" paths, preventing high-frequency gain realization.
    *   Overall, the circuit remains in the "low gain - low bandwidth" region, far from the optimized target in dynamic response.

3.  **Power Consumption**: The current power consumption is 11.15 μW, still below the target of 17.77 μW, leaving about 6.62 μW of usable margin. Although slightly higher than the last round (10.87 μW), it is far from being fully utilized. In its structure, the sizes of `w3~w4` (each 9.0u), `w5~w6` (each 8.7u), `w13~w14` (each 4.0u) have been increased, but **their ratio to total current** (currently about 30-35 μA) has not reached the optimal configuration. This provides "physical space" for adding more size but must be used cautiously to avoid exceeding limits.

4.  **Phase Margin**: The current PM is 88.83°, slightly lower than the last round (88.98°), but still far above the target of 70.95°, indicating good overall circuit stability. However, this high margin is not driven by "stable design" but by its "low bandwidth"-meaning there is a large distance between the dominant and second poles, but the bandwidth itself is insufficient. This is a typical manifestation of **high margin and low frequency**, a kind of "false high," not representing high performance. Once bandwidth increases, this margin may decrease, so optimization must **simultaneously ensure margin and frequency**, avoiding its "collapse."

5.  **Key Transistor Analysis**:
    *   **w3~w4 (NMOS input transistors)**: Adjusted to 9.0u, their current possible "peak," but there is still room for improvement. Their current (about 150 μA) is still insufficient under high current.
    *   **w5~w6 (PMOS input transistors)**: Adjusted to 8.7u, maintaining a good proportion with w3~w4, but not yet matching the ideal current-transconductance value.
    *   **w1~w2 (tail current)**: Set to 1.0u, still too small, one reason for its low "mode" (modulus).
    *   **w13~w14 (output stage)**: Set to 4.0u, key for driving the output, but currently only at "minimum drive" level, difficult to support high gain and bandwidth.
    *   **w7~w10 (bias current mirrors)**: Their size is 1.5u, need to maintain stability, but minor adjustments can be considered.
    *   **w15~w17 (auxiliary bias)**: Their values (1.5u, 1.4u, 0.9u) are set reasonably, but too small.
    *   **w16~w17**: The combination of 1.4u and 0.9u is slightly imbalanced but not a major issue.
    *   **w18~w19**: As matching transistors, their size is small (1.7u), used for current mirrors, but need to match `w13~w14`, which they currently do not.
    *   **Overall Current Structure**: Total current is about 11.15 μW, and the **gain and bandwidth per unit power** are still low, indicating poor **"cost-efficiency" (Efficiency)**.

6.  **Optimization Direction Summary**:
    *   **Most Pressing Task**: **Significantly increase gain** (+9.06 dB) and **GBW** (+412 kHz) without letting power exceed 17.77 μW.
    *   **Key Optimization Paths**:
        *   Further increase `w3~w4` and `w5~w6` sizes (can try 9.0u → 9.0u).
        *   Simultaneously, increase `w13~w14` (currently 4.0u) to enhance drive capability.
        *   Optimize the coordination of `w1~w2` and `w12~w13`, increase their current to support stronger transconductance.
        *   Ensure "reasonable" settings for auxiliary transistors like `w15~w17` to prevent "current theft" or being too small.
    *   **Rational Use of Power Margin**: Current total power is 11.15 μW, still far below 17.77 μW, with 6.62 μW available. Under reasonable allocation, `w3~w6` can be increased, and `w13~w14` strengthened to promote gm improvement.
    *   **Control and Matching**: Avoid "sudden increases" in some transistor sizes that destroy matching; their "reverse" and "mirror" parameters (e.g., w1~w2, w7~w8) must be adjusted synchronously.
    *   **Note the Limit Point**: Pushing `w3~w4` toward 9.0u is already the current limit but still below the ideal value. Need to match with `w5~w6` and then systematically increase.
    *   **Introduce "Quasi-Matching" Thinking**: Current parameters are still loose; **all related transistors (w3, w4, w5, w6, w13, w14)** must be adjusted with a coordinated strategy.

In summary, although the current design has taken small steps forward, its key performance metrics (gain and bandwidth) are still worlds apart from the target. **A larger "increase" of main current sources and input pairs is needed**, while "rationally allocating" power consumption, ensuring "leapfrog" growth without exceeding the limit. If precise parameter tuning in the next round can be achieved, it is expected to approach or reach the target within the limited number of calls.
</Analyze>
<tool_call>
Function: acs_simulate
Arguments: {"variables": {"w1": "1.0u", "w2": "1.0u", "w3": "9.0u", "w4": "9.0u", "w5": "9.0u", "w6": "9.0u", "w7": "1.5u", "w8": "1.2u", "w9": "0.9u", "w10": "1.5u", "w11": "1.5u", "w12": "1.0u", "w13": "4.5u", "w14": "4.5u", "w15": "1.5u", "w16": "1.5u", "w17": "0.9u", "w18": "1.7u", "w19": "1.7u"}}
</tool_call>
<tool_response>
Simulation Result:
{'gain': '70.3525', 'gbw': '199750', 'pw': '1.15578e-05', 'pm': '88.6357'}
</tool_response>

    \end{lstlisting}
    \end{tcolorbox}
    \caption{An example of LLM response: Turn 4.}
    \label{fig:example_response_turn_4}
\end{figure}

\begin{figure}[htbp]
    \centering
    \begin{tcolorbox}[
        enhanced,
        colframe=deepblue,
        colback=lightblue!10,
        boxrule=1pt,
        arc=2mm,
        title={Example LLM Response Turn 5 (score=0.52)},
        fonttitle=\bfseries,
        top=0pt,
        bottom=0pt,
        left=0pt,
        right=0pt
    ]
    \begin{lstlisting}[basicstyle=\ttfamily\scriptsize] 
<Analyze>
According to the simulation results after this parameter adjustment, the circuit performance has made **encouraging but still insufficient** progress. It has advanced "leapfrog" from the initial state but is **still far from the performance targets**. The comprehensive analysis is as follows:

1.  **Gain**:
... (Due to space limitations, we omitted some reasoning parts in paper.)

5.  **Key Transistor Analysis**:
    *   **w3~w4 (NMOS input pair)**: Both 9.0u, the maximum they can currently "support." Their transconductance has accumulated to some extent but is still insufficient compared to the ideal value.
    *   **w5~w6 (PMOS input pair)**: Also both 9.0u, well "matched," but their transconductance is still low.
    *   **w1~w2 (tail current)**: Set to 1.0u, its "small" value results in insufficient current for the entire differential pair, and this value is only for biasing, not for "amplification."
    *   **w13~w14 (output stage transistors)**: Now 4.5u, the core of output drive. Their current value is far below the level they should have under 17.77 μW (proportionally estimated, can be increased to above 5.0u).
    *   **w1~w12, w15~w19**: All settings are relatively small but basically "adequate" for their functions.
    *   **Overall Current Distribution**: Total current is about 11.56 μW, while **current effectively used for gain** (mainly from w3~w6) accounts for less than 30 μA. If `w3~w6` can be increased to 9.0u, and `w13~w14` and `w1~w2` are reasonably enlarged, the "improvement" potential is huge.

6.  **Optimization Direction Summary**:
    *   **Core Objective**: Take gain (+~8.8 dB) and GBW (+~390 kHz) as the "core" for this and subsequent optimizations.
    *   **Main Methods**:
        *   Since `w3~w6` can still be increased (currently 9.0u), without excessively increasing power consumption, they can be **further enlarged** to the limit of 9.0u, or even try larger widths.
        *   Simultaneously, **push `w13~w14`** (currently 4.5u) toward 4.7-5.0u, which can significantly improve their drive and transconductance.
        *   Utilize the remaining 6.21 μW power margin to **precisely and reasonably increase main current paths**, while avoiding "branch" devices (e.g., w1~w2, w8~w11) "stealing current."
        *   Optimize "current mirror" matching: e.g., `w10` and `w11` are both 1.5u; fine-tuning to support the main path.
        *   **Key Point**: Based on `w5~w6` and `w3~w4` both being 9.0u, their "impedance" (from physical size) and "transconductance" (from current) have not reached the ideal state. Increasing `w13~w14` and appropriately increasing `w1~w2` can achieve a "clever" current boost.
    *   **Systematic Thinking**: The goal is not "stacking sizes" but "precise allocation." The current "signal-current-impedance-capacitance" has not formed the optimal combination. A "leapfrog" parameter call is needed to "jump" to a position closer to the target.
    *   **Reasonable Estimation**: Based on current circuit parameters, if `w3~w4` and `w5~w6` are adjusted to 9.0u (already maximum), and `w13~w14` optimized to 4.8-5.0u, with their ratios improved while keeping `w1~w2` at 1.0u, gain is expected to break through 74 dB. Meanwhile, if `w13~w14` and `w16~w17` (whose current greatly affects efficiency) are well optimized, coupled with the expansion of `w5~w6`, **significant breakthroughs are expected in the next call**.
    *   **Risks and Opportunities**: If a better way to combine `w3~w6` and `w13~w14` can be found, performance can achieve a "three-stage jump." As long as power remains within 17.77 μW, a leap is inevitable.

In summary, although this round of optimization has made slight progress, the gap remains huge. **As long as core transistors (w3~w6) and driver transistors (w13~w14) are further, reasonably, and systematically enlarged, while utilizing the remaining power margin for precise adjustments, it is expected to achieve "crossing the line" performance breakthrough in the next call**, greatly approaching the target.
</Analyze>
<tool_call>
Function: acs_simulate
Arguments: {"variables": {"w1": "1.0u", "w2": "1.0u", "w3": "9.0u", "w4": "9.0u", "w5": "9.0u", "w6": "9.0u", "w7": "1.5u", "w8": "1.3u", "w9": "0.9u", "w10": "1.5u", "w11": "1.5u", "w12": "1.0u", "w13": "5.0u", "w14": "5.0u", "w15": "1.5u", "w16": "1.5u", "w17": "0.9u", "w18": "1.7u", "w19": "1.7u"}}
</tool_call>
<tool_response>
Simulation Result:
{'gain': '70.6467', 'gbw': '220433', 'pw': '1.18618e-05', 'pm': '88.4491'}
    \end{lstlisting}
    \end{tcolorbox}
    \caption{An example of LLM response: Turn 5.}
    \label{fig:example_response_turn_5}
\end{figure}



\end{document}